\documentclass{article}
\usepackage[nonatbib,preprint]{neurips_2022}
\usepackage{url}            

\author{%
Juan Zuluaga-Gomez$^{\star,1,2}$ \quad Karel Vesel\'{y}$^{3}$ \quad Igor Szöke$^{3,4}$ 
\quad Alexander Blatt$^{5}$
\quad Petr Motlicek$^{1,3}$ \quad \\
\textbf{Martin Kocour}$^{3}$ \quad \textbf{Mickael Rigault}$^{6}$ \quad \textbf{Khalid Choukri}$^{6}$ \textbf{Amrutha Prasad}$^{1,3}$ \\
\textbf{Seyyed Saeed Sarfjoo}$^{1}$ \quad \textbf{Iuliia Nigmatulina}$^{1}$ \quad \textbf{Claudia Cevenini}$^{7}$ \\
\textbf{Pavel Kol\v{c}árek}$^{8}$ \quad \textbf{Allan Tart}$^{9}$ \quad 
\textbf{Jan \v{C}ernocký}$^{3}$ \quad \textbf{Dietrich Klakow}$^{5}$ \\
$^{1}$Idiap Research Institute, Martigny, Switzerland \\
$^{2}$Ecole Polytechnique Federale de Lausanne (EPFL), Lausanne, Switzerland \\ 
$^{3}$Brno University of Technology, Speech@FIT, IT4I CoE, Brno, Czech Republic \\
$^{4}$ReplayWell, Brno, Czech Republic \\ 
$^{5}$Saarland University, Saarland Informatics Campus, Germany \\ 
$^{6}$Evaluations and Language Resources Distribution Agency (ELDA), Paris, France \\
$^{7}$Romagna Tech, Forli, Italy \\ 
$^{8}$Honeywell, Brno, Czech Republic \\ 
$^{9}$OpenSky Network, Burgdorf, Switzerland \\
$^{\star}$Corresponding author: \textbf{juan-pablo.zuluaga@idiap.ch}
}

\usepackage[utf8]{inputenc} 
\usepackage[T1]{fontenc}    
\usepackage[hidelinks]{hyperref}       
\usepackage{booktabs}       
\usepackage{amsfonts}       
\usepackage{amsmath}
\usepackage{nicefrac}       
\usepackage{microtype}      
\usepackage{wrapfig}
\usepackage{multirow}
\usepackage{subcaption}

\usepackage{pdfpages}

\usepackage[numbers]{natbib}
\usepackage{bbding}
\definecolor{Gray}{gray}{0.9}

\newcommand{\argmax}{\operatornamewithlimits{argmax}}
\newcommand{\func}[1]{\mathrm{#1}}
\newcommand{\V}[1]{\mathbf{#1}}

\usepackage{color, colortbl}
\usepackage{listings}
\definecolor{codegreen}{rgb}{0,0.6,0}
\definecolor{codegray}{rgb}{0.5,0.5,0.5}
\definecolor{codepurple}{rgb}{0.58,0,0.82}
\definecolor{backcolour}{rgb}{0.95,0.95,0.92}

\lstdefinestyle{mystyle}{
    language=XML,
    backgroundcolor=\color{backcolour},   
    commentstyle=\color{codegreen},
    keywordstyle=\color{magenta},
    numberstyle=\tiny\color{codegray},
    stringstyle=\color{codepurple},
    basicstyle=\ttfamily\footnotesize,
    breakatwhitespace=false,         
    breaklines=true,                 
    captionpos=b,                    
    keepspaces=true,                 
    numbers=left,                    
    numbersep=5pt,                  
    showspaces=false,                
    showstringspaces=false,
    showtabs=false,                  
    tabsize=2,
    morekeywords={segment,data,version,type,release},    
}
\title{ATCO2 corpus\\ A Large-Scale Dataset for Research on Automatic Speech Recognition and Natural Language Understanding of Air Traffic Control Communications}

\begin{document}
\maketitle

\begin{abstract}
\textbf{Abstract:} Personal assistants, automatic speech recognizers and dialogue understanding systems are becoming more critical in our interconnected digital world. A clear example is air traffic control (ATC) communications. 
ATC aims at guiding aircraft and controlling the airspace in a safe and optimal manner.
These voice-based dialogues are carried between an air traffic controller (ATCO) and pilots via very-high frequency radio channels.
In order to incorporate these novel technologies into ATC, large-scale annotated datasets are required to develop the data-driven AI systems. Two examples are automatic speech recognition (ASR) and natural language understanding (NLU). 
However, ATC is considered a low-resource domain. 
In this paper, we make several contributions aiming at overcoming these disadvantages. First, we introduce the \textit{ATCO2 corpus}, a dataset that aims at fostering research on the challenging ATC field, which has lagged behind due to lack of annotated data. Second, we open-source a GitHub repository\footnote{Our code will be stored in the following public GitHub repository \url{https://github.com/idiap/atco2-corpus}.} that contains data preparation and training scripts useful to replicate some of our baselines related to ASR and NLU for ATC communications. 
The \textit{ATCO2 corpus} covers 1) data collection and pre-processing, 2) pseudo-annotations of speech data, and 3) extraction of ATC-related named entities.
The \textit{ATCO2 corpus} is split into three subsets. 1) \textit{ATCO2-test-set corpus} contains 4 hours of ATC speech with manual transcripts and a subset with gold annotations for named-entity recognition (callsign, command, value).
2) The \textit{ATCO2-PL-set corpus} consists of 5281 hours of unlabeled ATC data enriched with automatic transcripts from an in-domain speech recognizer, contextual information (list of relevant n-gram sequences per utterance), speaker turn information, signal-to-noise ratio estimate and English language detection score per sample. These two are available for purchase through ELDA in \url{http://catalog.elra.info/en-us/repository/browse/ELRA-S0484/}. 3) The \textit{ATCO2-test-set-1h corpus} is a one-hour subset from the original test set corpus, that we are offering for free in the following website: \url{https://www.atco2.org/data}. 
We expect the \textit{ATCO2 corpus} will foster research on robust ASR and NLU not only in the field of ATC communications but also in the general research community. 
\end{abstract}

\textbf{Keywords}: Robust Automatic Speech Recognition, Natural Language Processing, Air Traffic Control Communications, Spoken Language Understanding, Signal Processing

\section{Introduction}
\label{sec:introduction}

The corpus introduced in this research is within the domain of civil air traffic control (ATC) communications and management. ATC aims at managing the airspace in a safe and optimal manner. The communication is either via spoken or data-link messages, while the time-critical messages are always spoken. These communications involve an air traffic controller (from now on, ATCO) issuing spoken flight instructions to aircraft pilots during all phases of the flight. The dialogue follows a well-defined grammar and set of rules that ensures safety, reliability, and efficiency~\cite{allclear,kocour2021automatic}. This can be seen as a multi-speaker and multi-turn conversation.

Commonly, an ATCO addresses several pilots in a short period of time, which in turns becomes the main cause of increased workload and limiting factor to increase the overall system capacity, i.e., there is large space for optimization by only reducing ATCO's workload. A significant bottleneck in the pipeline is the considerable latency arising from an ATCO issuing a command by voice and inserting it manually into the ATCO's workstation (for control and record). 
Recent advances in automatic speech recognition (ASR) and natural language processing (NLP) technologies have opened new ways where ATCO's workload can be reduced\footnote{In fact, workload reduction is also translated in reduced flight time, which decreases overall operational costs and the environmental impact of aircraft.} by integrating different systems in a cascade fashion. The systems for extracting the actual meaning from the original audio signal are commonly known as spoken language understanding (SLU).

Our previous works have made sizeable progress on independent systems for ATC, such as, robust ASR~\cite{zuluagagomez20_interspeech}, NLP~\cite{zuluaga2020automatic}, and diarization and segmentation~\cite{kocour2021automatic}.
However, until today, these systems are close to non-existent in real-life ATC operations. In part, this is due to the intrinsic complexity of the task, and mainly to the lack of annotated data available to train these data-driven systems~\cite{cordero2012automated}. 

This paper introduces the \textit{ATCO2 Corpus} derived from a joint contribution from Clean Sky 2 Joint Undertaking (JU) and EU-H2020. ATCO2\footnote{\textbf{A}u\textbf{T}omatic \textbf{CO}llection and processing of voice data from \textbf{A}ir-\textbf{T}raffic \textbf{CO}mmunications, website: \url{https://www.atco2.org/}.} project developed a platform to collect, organize, pre-process and automatically annotate ATC dialogues\footnote{We believe this pipeline can be easily adapted to other applications, where data scarcity is a latent problem, but access to unlabeled/non-annotated data is permissible e.g., patient–physician dialogues.}. The main bottleneck towards ASR or natural language understanding (NLU) techniques for ATC are the lack of annotated data. Further, its collection and annotation requires trained personal, thus, the whole pipeline becomes excessively costly and impractical.
This study presents how the entire data collection and annotation process can be efficiently accelerated by using already existing machine learning (ML) concepts.

\begin{figure}[t]
    \centering
    \includegraphics[width=0.8\linewidth]{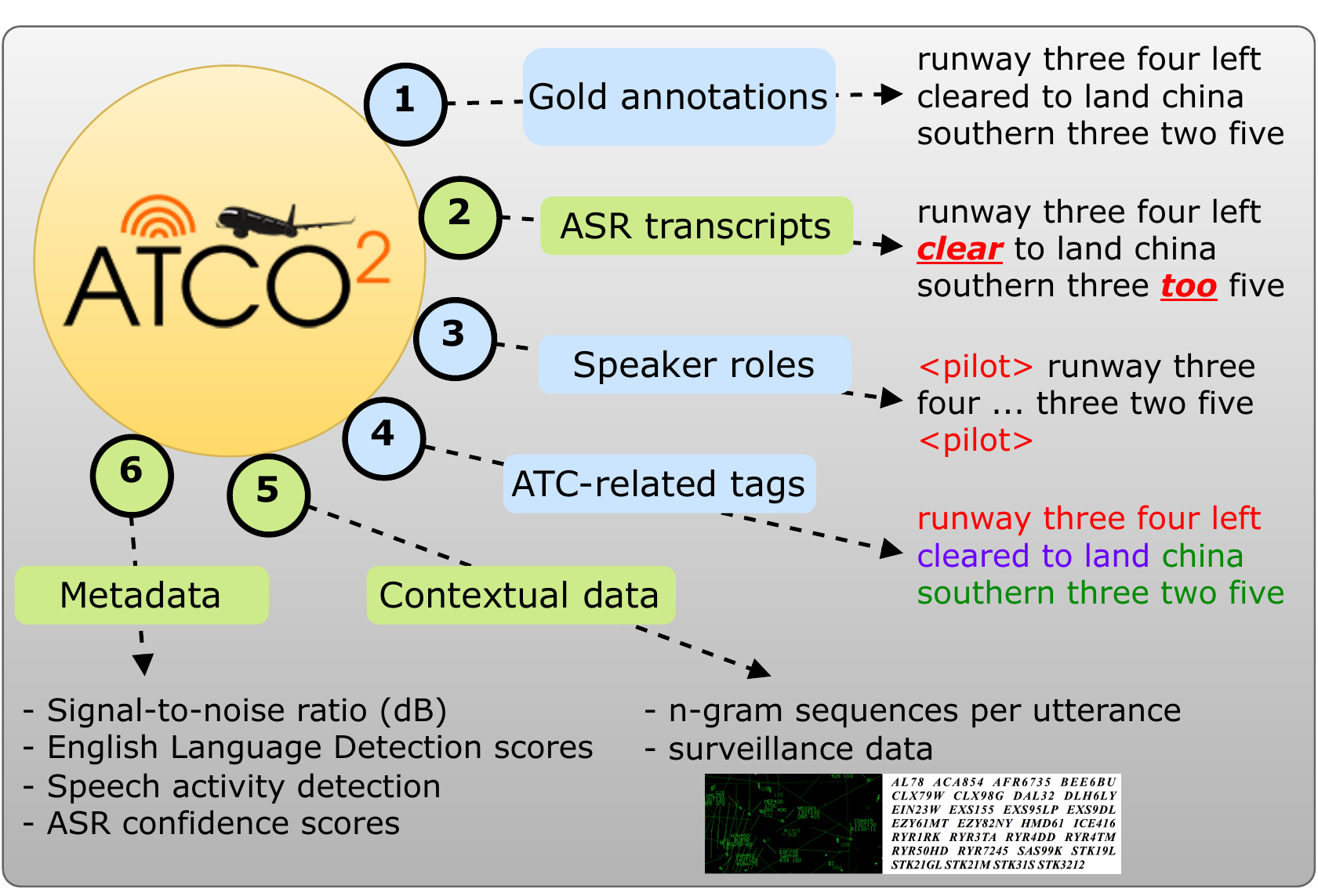}
    \caption{\textbf{\textit{ATCO2 corpus} ecosystem}. Blue circles denote annotations only available for \textit{ATCO2 test set corpus}. Green circles denote annotations and metadata available for both \textit{ATCO2 test set} and \textit{ATCO2 pseudo-labeled} corpus sets (see Table~\ref{tab:databases} bottom).}
    \label{fig:atco2-corpus-information}
\end{figure}

The overall \textit{ATCO2 Corpus} ecosystem is depicted in Figure~\ref{fig:atco2-corpus-information}. We release two corpora targeted to ATC for research in robust ASR, NLP and NLU, i.e., the i) \textit{ATCO2-test-set corpus} and ii) \textit{ATCO2 pseudo-labeled set corpus} (\textit{ATCO2-PL-set corpus}). The former contains word-level and named entities\footnote{Our Named Entity Recognition (NER) classes are: Callsign, Command, Value and Unnamed Phrase. The NER labels can be used to train/test an SLU system for slot filling.} gold annotations. In total, we release 4 hours of speech with various useful metadata (see the blue circles from Figure~\ref{fig:atco2-corpus-information}). The latter, \textit{ATCO2-PL-set corpus}, contains $\sim$5281\,hours of ATC audio recordings, where each utterance includes a detailed set of metadata. For instance: pseudo-transcripts obtained from an in-domain ATC ASR system (the pseudo transcripts contain also diarization and segmentation information), contextual information (list of word sequences for lattice-boosting of callsigns), signal-to-noise ratio (SNR) estimate, and English language detection (ELD) scores (see the green circles from Figure~\ref{fig:atco2-corpus-information}).
Even though this is not the first publicly available corpus related to ATC communications~\cite{N4NATO,ATCOSIM,UWB_ATCC,LDC_ATCC}, to author's knowledge, this is the first corpus that conveys annotated data for text and spoken-based tasks e.g., named entity recognition (NER), slot filling (SF), or sequence classification.

\begin{figure}[t]
    \centering
    \includegraphics[width=0.8\linewidth]{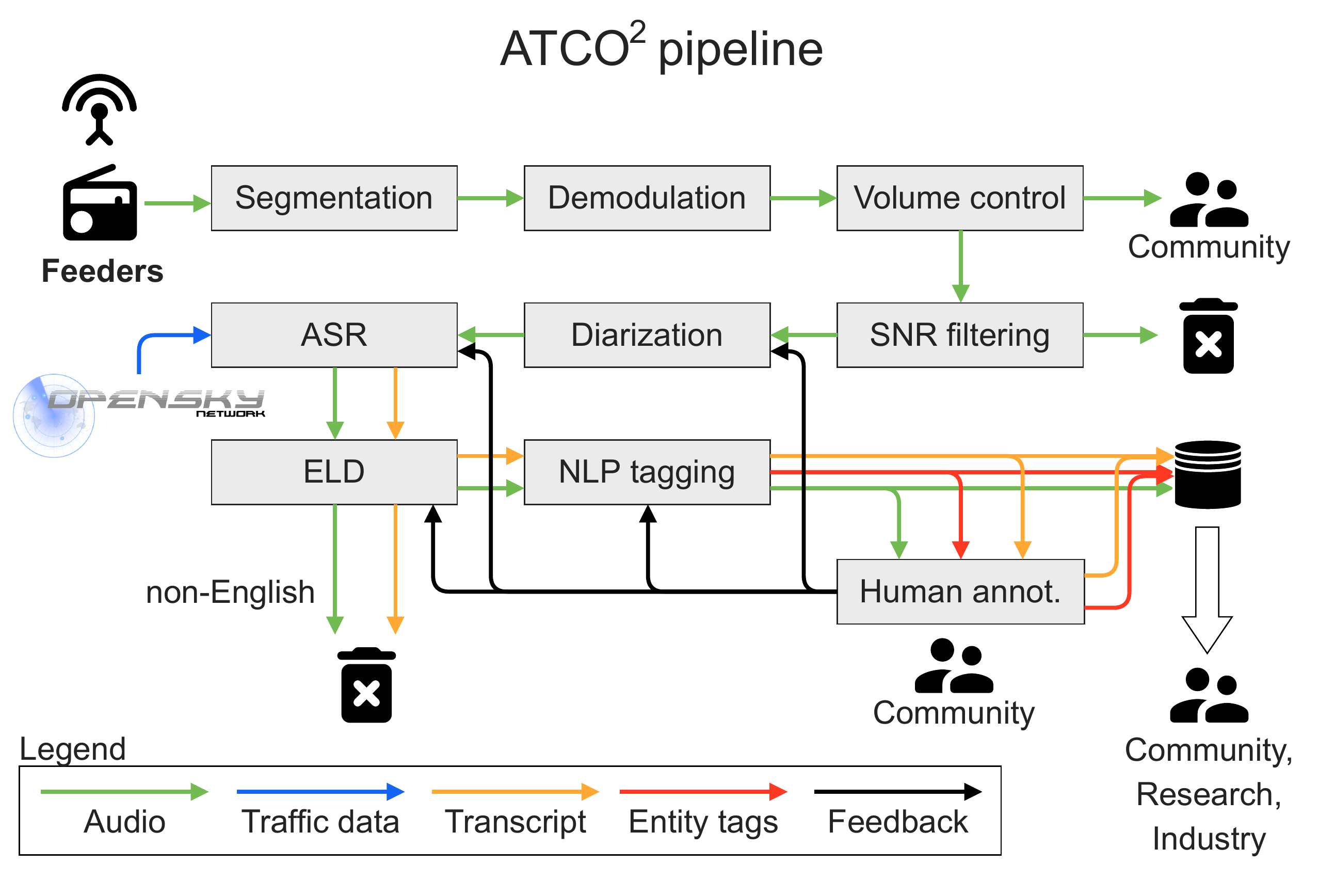}
    \caption{Data collection and data-processing pipeline developed in ATCO2 project.}
    \label{fig:pipeline}
\end{figure}

An overview of the data processing pipeline developed by ATCO2 and used to collect the \textit{ATCO2 corpus} is depicted in Figure~\ref{fig:pipeline} (a more detailed description is in Section \ref{subsec:data-processing-pipeline}). 
The data processing pipeline consists of 1) speech pre-processing tools (segmentation, volume adjustment and discarding noisy recordings), 2) diarization (split audio per speaker), 3) ASR, 4) English language detection (ELD), 5) speaker role detection (SRD) e.g., ATCO or pilot, and 6) labeling of callsigns, commands and values with named entity recognition (NER). We used this pipeline to produce the \textit{ATCO2-PL-set corpus} and \textit{ATCO2-test-set corpus}, which cover more than ten airports worldwide. The ATCO2 corpus is publicly available in ELDA catalog at the following URL: \url{http://catalog.elra.info/en-us/repository/browse/ELRA-S0484/}. Further details about the data collection and pre-processing is covered in the Appendix~\ref{appendix:full-pipeline} and our previous paper~\cite{kocour2021automatic}.

We also emphasize that the developed pipeline was not only used to collect the data presented in this paper, but also is running live at \url{https://www.spokendata.com/atco2}. The data is automatically fed, filtered, and pre-transcribed, so the amount of data will increase. We are searching for volunteers, both for feeding data from new airports (see Section \ref{subsec:feeders}) or for correcting the automatic transcripts (see Sections \ref{subsec:transcriber}). In general, the \textit{ATCO2 corpus} can be used to produce a robust ASR system for the ATC domain, and with its NER annotations it is possible to train models for SLU applications for extracting meaning from speech (e.g., callsign or command detection). However, we also believe that the pipeline developed by ATCO2 can be also adapted to data collection and annotation of different domains, e.g., call-centers conversations, or medical recordings.

\textbf{Motivation:} speech and text-based processing tools for ATC data could work better if we had a large amount of reliably annotated data. However, the collection and manual annotation requires qualified personnel, and it is costly. In addition, the recordings are often noisy (SNR below 15 dB), accented or with high speech rate (compared to conversational, read or spontaneous speech). Aligned to solve this, \textit{ATCO2 corpus} answers four big challenges:

\textbf{Current ATC corpora are limited to automatic speech recognition}. However, ASR is only a small submodule of the whole pipeline and many more downstream tasks are indeed required, e.g., ATC-related NER or callsign detection and extraction. In our case, those are callsigns, commands and their values. \textit{ATCO2 corpora} goes in this line and further by releasing gold annotations to train systems on ASR, NER, ELD, and SRD.

\textbf{Research on ATC communications has lagged behind due to the lack of annotated data}. 
The primary rational motive is the high annotation cost. ATC communications require eight to ten man-hours effort~\cite{cordero2012automated} to annotate one hour of raw controller-pilot dialogues. Primarily because it requires highly trained participants, often active or retired ATCOs. 
In total, after further pre-processing (e.g., silence and noisy segments removal) around one man-week work yields roughly an hour of annotations without silences~\cite{cordero2012automated,ferreiros2012speech}.
This number increases if further metadata is required, e.g., word-level tagging for NER. We address these issues by developing an efficient pipeline to collect, pre-process and automatically annotate ATC data (depicted in Figure~\ref{fig:pipeline}). Using pre-transcribed data rather than transcribing from scratch reduced drastically the whole annotation process time period. Similarly, our observations during ATCO2 project revealed that the real-time factor (RTF) among the data transcribers varied drastically. For instance, untrained transcribers exhibited up to 50 RTF for transcribing ATC speech, including channel and NER tagging. However, trained transcribers reached as low as 20 RTF for the whole transcription process.

\textbf{Domain shift between ATC and non-ATC corpora is too strong}.
Current ATC corpora contain data from only a few airports, and some were collected in clean and quiet simulation or training rooms~\cite{N4NATO,ATCOSIM,UWB_ATCC}. 
Even though ATC speech should follow the same phraseology, the data from different airports substantially differ due to local conventions, speakers accent and rate of speech. All this together creates a considerable domain shift.
Current ASR engines on the ATC domain are tailored to a particular airport\footnote{Other EU-funded projects, like \href{http://www.malorca-project.de/wp}{Malorca} or \href{https://www.haawaii.de/wp/}{HAAWAII}, only focus on developing ASR tools for one or at most two airports per project.}. 
Our ambition was, however, to collect and release annotated and pseudo-labeled recordings from many airports, which in turns can foster the training of more airport-agnostic ASR, NLU, and SLU systems. Moreover, data from non-ATC datasets like LibriSpeech\footnote{This also includes other popular corpora, such as, CommonVoice~\cite{ardila2019common} or Switchboard~\cite{godfrey1992switchboard}.}~\cite{panayotov_librispeech_ICASSP2015} do not match the ATC acoustics and its use does not help in the ASR training~\cite{zuluagagomez20_interspeech}.

\textbf{Applicability on general spoken language understanding}. Even though the \textit{ATCO2 corpus} and baselines presented here are aimed at a niche application (air traffic control communications), we believe that general-purpose research on NLU/SLU can widely benefit from this corpus. Most of the current benchmarks on SLU are widely saturated, where the performances (e.g., F1-scores) are near perfect, a couple of examples are ATIS~\cite{hemphill1990atis} or SNIPS~\cite{coucke2018snips} datasets. 
Differently, \textit{ATCO2 corpus} is composed of very noisy voice recordings (often below 15 dB SNR). The audio data is collected from devices (see Figure~\ref{fig:atco-receiver}) owned by a community of volunteers (see Section~\ref{subsec:feeders}), thus, it is more natural to find noisy data. This, in turn, increases the challenge of standard ASR systems, e.g., WERs of $\sim$30\% or above (see our previous baselines~\cite{zuluaga2022does,kocour2021automatic}). 
We hope that the research community will build upon the \textit{ATCO2 corpus} presented in this paper. Additionally, we hope that the presented baselines will foster research in the fields of ASR and NLP for ATC communications.

The paper is organized as follows.
Section 2 covers previous work on ASR and NLP directed to ATC communications.
Section 3 explains our proposed methodologies for the standardization of ATC communications annotation process. The data collection protocol, pre- and post-processing steps undertaken during the annotation process of \textit{ATCO2 corpora} are described in Section 4.
\textit{ATCO2 corpora} data statics are reviewed in Section 5. Section 6 and 7 convey the proposed baselines on ASR and NLP, respectively. Section 8 covers the main legal and ethical implications of ATC data collection. Finally, we conclude this paper in Section 9 with final remarks and prospect of future work.


\section{Previous Work on ATC Corpora}

Currently, there is a huge diversity of databases related to speech and text tasks that have been promoting advances in artificial intelligence (AI). However, ATC communications are still considered an under resourced and underexplored area
\cite{zuluaga2022does,zuluaga2021bertraffic}. Despite the growing interest in text and speech technologies for ATC, there is not a commercial ASR engine due to: (i)~deficiency in terms of required performance (under 5\% WER~\cite{ohneiser2021robust}), and (ii)~lack of large-scale annotated speech data. The costly data collection and annotation makes it impractical, when transfer to a new airport requires data collection and annotation. 

\subsection{Background}

Research seeking to aid ATCOs by ASR date as back as late 70s'.~\cite{beek1977} proposed a system for isolated word recognition, speaker verification and commands recognition for military applications.
Exploratory research towards integration of ASR technologies to aid ATCOs started in the late 80s with Hamel et al.~\cite{hamel1989}. 
Several other research directions cover user-friendly and robust automatic systems to train ATCOs, or the so called `pseudo-pilots'~\cite{matrouf1990adapting}. Akin training systems have been proposed in~\cite{matrouf1990adapting,tarakan2008automated,ferreiros2012speech}. We shortlist the three biggest European-based projects that aim at developing speech and text-based tools to aid ATCOs in their daily tasks. 
Initially, MALORCA\footnote{MAchine Learning Of speech Recognition models for Controller Assistance, website: \url{http://www.malorca-project.de/wp/}.} project was a step forward in demonstrating that ASR tools can cut down ATCOs workload~\cite{helmke2016reducing} while increasing the overall efficiency~\cite{helmke2017increasing}. 
Then, HAAWAII\footnote{Highly Automated Air traffic controller Workstations with Artificial Intelligence Integration, website: \url{https://www.haawaii.de/wp/}.} project has led initiatives to extract key entities (e.g., NER or SF) in the transcribed dialogues produced by an ASR system~\cite{kleinert2021automated}.
Finally, ATCO2 project (our corpora) aimed at reducing the human work needed to develop ASR and SLU tools for ATC, mainly by integrating semi-supervised techniques in these systems~\cite{kocour2021automatic,zuluaga2020automatic}. While the MALORCA and HAAWAII corpora are not public, in ATCO2 we developed a pipeline to collect large quantities of ATC speech data, which are distributed to the public through ELDA.

\subsection{Command-related ATC Corpora vs Standard Corpora}

The ATC speech corpora differ vastly from the standard ASR-training corpora.
The root of the discrepancy is not only in grammar and vocabulary, but also in the audio quality. 
The standard corpora like Librispeech~\cite{panayotov_librispeech_ICASSP2015}, Common Voice~\cite{ardila2019common}, AMI\,\cite{mccowan2005ami} or TED-LIUM~\cite{rousseau2012ted} either target conversational, read or spontaneous speech while also being mostly regarded as `clean speech'. On the other hand, ATC corpora comprises considerably higher noise levels e.g., normally below 15dB signal-to-noise (SNR) ratio, heavily accented, high speech rate and artifacts. 
Previous work has demonstrated that the use of standard corpora do not bring significant improvement in ASR for ATC~\cite{zuluaga2020automatic}.

Even though, ATC English corpora share common grammar, there is still a domain shift caused by non-native speakers. One example are ATCOs from Switzerland. Even though they are from the same country, accent varies depending on the location. This, in turn, increases the challenge of developing robust enough systems that generalize well across different in-domain environments. Therefore, a non-adapted ASR or NLP system will provide significantly worse performance due to unseen accents, out-of-vocabulary (OOV) words or simply due to discrepancy in the recording procedure.

Further details about the corpora produced by previous projects related to ATC communications are covered in Table~\ref{tab:databases}. Current ATC corpora can be classified into two, public and private databases. Public databases normally require a small fee and sometimes are restricted to only-research purposes. While private corpora\footnote{In fact, nearly all ongoing and former projects in the area of ATC prohibit the release of databases, code, and AI models due to privacy issues.} are only usable along the concerned project, for instance, to train and test their ATC-related systems. One example is MALORCA, where the two produced corpora, \textit{Prague} and \textit{Vienna} datasets, are widely used by partners from HAAWAII and ATCO2 projects.


\begin{table}[t]
    \caption{Air traffic control communications related databases. This table list public and private ATC databases. The \textit{ATCO2 corpora} are public databases. $^\dagger$full database after silence removal. $^{\dagger\dagger}$speaker accents depend on the airport's location, however, the accent of pilots are not known at any time of the communication due to privacy regulations.}
    \label{tab:databases}
    \centering

    \resizebox{0.99\textwidth}{!}{
    \begin{tabular}{l p{6cm} c p{2.5cm} ll} \toprule
        \multicolumn{1}{l}{\textbf{Database}} & \textbf{Details} &
        \textbf{Licensed} &
        \textbf{Accents} & \textbf{Hours}$^\dagger$ & \textbf{Ref} \\ 
        \midrule
        \rowcolor{Gray}
        \multicolumn{6}{c}{\textbf{\textit{Private databases}}} \\
        \midrule
        \quad HAAWAII & Real data from Iceland and London airports & \XSolidBrush & \raggedright Icelandic, British & 47 & \cite{zuluaga2022does}  \\
        \quad MALORCA & Real data from Vienna and Prague airports & \XSolidBrush & \raggedright German, Czech & 13 & \cite{kleinert2018semi,srinivasamurthy2017semi}  \\
        \quad AIRBUS & Real data from Toulouse-Blagnac airport & \XSolidBrush & French & 100 &  \cite{AIRBUS} \\
        \quad VOCALISE & Real data from terminal maneuvering area and area control center in France & \XSolidBrush & French & 150 &  \cite{graglia2005vocalise} \\
        \quad ENAC & Real data from two French en-route control centers and one major airport & \XSolidBrush & French & 22 &  \cite{lopez2013linguistic} \\
        \midrule
        \rowcolor{Gray}
        \multicolumn{6}{c}{\textbf{\textit{Public databases}}} \\
        \midrule
        \quad ATCOSIM & Simulated in studio, added cockpit noise. Recordings split by gender (Male/Female) & \checkmark & \raggedright Swiss~German, German, French & 10.7 &  \cite{ATCOSIM} \\
        \quad UWB-ATCC & Real data from Prague airport & \checkmark & Czech & 13.2 & \cite{UWB_ATCC}  \\
        \quad LDC-ATCC & Real data from 3 US airports: Logan International, Washington National and Dallas Fort Worth airports & \checkmark & American English & 26.2 & \cite{LDC_ATCC}  \\
        \quad HIWIRE & Simulated in studio, ATC prompts, added cockpit noise & \checkmark & French, Greek, Italian, Spanish & 28.7 &  \cite{HIWIRE} \\
        \midrule
        \rowcolor{Gray}
        \multicolumn{6}{c}{\textbf{\textit{Released corpora by ATCO2 project}}} \\
        \midrule
        \textbf{\textit{ATCO2 corpora}} & Data from different airports and countries & & Several$^{\dagger\dagger}$ & & \\
        
        \cline{2-6}\rule{0pt}{3ex}$\hookrightarrow$ \textit{ATCO2-test-set} & Real data for ASR and NLP research. & \checkmark & $\hookrightarrow$ & 4 &  \\

        $\hookrightarrow$ \textit{ATCO2-PL-set} & Pseudo-labeled real data for research in ASR and \mbox{NLU}.  & \checkmark & $\hookrightarrow$ & 5381 & \cite{kocour2021automatic} \\

        \cline{2-6}
        \rule{0pt}{3ex} & \multicolumn{5}{c}{\cellcolor{Gray}\textbf{\textit{Free access databases releseased by ATCO2 project}}} \\
        \cline{2-6}
        \rule{0pt}{3ex}$\hookrightarrow$ \textit{ATCO2-test-set-1h} & 'ASR dataset': public 1 hour sample, subset of \textit{ATCO2-test-set-4h}. \url{https://www.atco2.org/data} & \checkmark & $\hookrightarrow$ & 1 & \cite{kocour2021automatic} \\

        $\hookrightarrow$ \textit{ATCO2-ELD set} & 'LID dataset': public dataset for English language detection. \url{https://www.atco2.org/data} & \checkmark & $\hookrightarrow$ & 26.5 & \cite{szoke21_interspeech} \\
        
        \bottomrule
    \end{tabular}
    }
\end{table}

\section{How To Transcribe Air Traffic Control Audio Data?}

This section reviews our collective efforts to provide an unambiguous and clear protocol on how to annotate ATC speech data. We aim at avoiding as much as possible the errors caused by OOV words and phonetic dissimilarities (e.g., ``hold in position`` and ``holding position", or, ``climb to two thousand" and ``climb two two thousand"). We also rely on the International Civil Aviation Organization (ICAO), which defines a standard phraseology~\cite{allclear} to reduce these errors during the communications.\footnote{Previous work in~\cite{helmke2018ontology} proposes a novel ontology agreed by several European institutions to annotate unambiguously these dialogues.} This section first formulates an approach to unify transcripts from different public ATC databases (see Table~\ref{tab:databases} and Appendix~\ref{appendix:unification}). Second, it discusses how to craft a lexicon tailored to ATC communications.

\subsection{Unification of Transcripts}

Differently from other corpora, ATC is regulated by a set of rules and a defined grammar. We have seen that in the already available databases (see Table~\ref{tab:databases}), the transcription process and annotation rules widely diverge. 
There is not a clear path to follow when it comes to data collection and annotation. 
Even for a single database, it can be challenging to specify and follow their transcription conventions.
One example are numbers. In ATC communications, numbers are key for addressing the aircraft or obtaining its speed or altitude. Several databases have opted to annotate numbers as digits (e.g., 1 $\rightarrow$ \textit{1}), while other databases have chosen to use words (e.g., 1 $\rightarrow$ \textit{one}). 
That is why the ATCO2 corpus also aims at providing a set of good practices and rules to correctly and unambiguously annotate ATC dialogues\footnote{Some of these rules and lessons can be easily adapted to other databases in the domain of ASR or NLU.}. Therefore, if we succeed in reducing the variability of “writing the same thing in many ways”, we can considerably reduce the errors committed by subsequent systems in the pipeline, e.g., ASR or NLU. 
The next logic step, before starting the annotation process, is to define a set of rules to either, unify the transcripts of already available corpora\footnote{We use these rules to normalize the transcripts of \textit{UWB-ATCC} and \textit{LDC-ATCC} databases (see Table~\ref{tab:databases} and~\ref{tab:databases_experiments}) for experimentation.} or, to annotate a new corpus. 

\begin{figure}[t]
    \centering
    \includegraphics[width=0.7\linewidth]{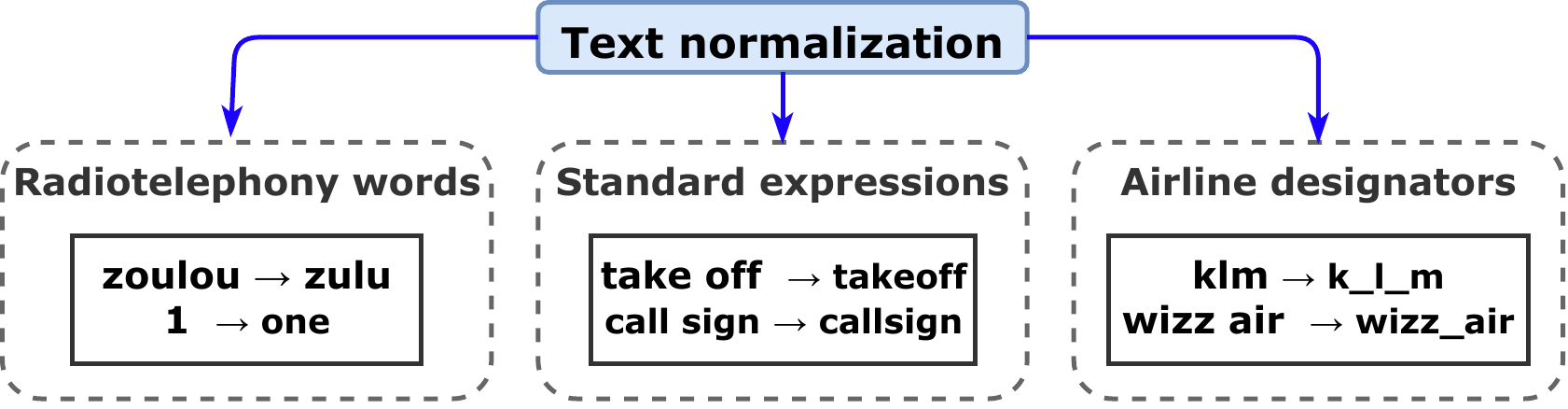}
    \caption{Text normalization applied to the transcription process of ATCO2 corpora. Further examples in Appendix~\ref{appendix:unification}.}
    \label{fig:text-normalization}
\end{figure}

In general, we apply three different text normalization approaches (see Figure~\ref{fig:text-normalization}) to foster good practices on the ATC-transcription process. We redirect the reader to Appendix~\ref{appendix:unification}, where additional mapping rules are covered in Table~\ref{tab:appendix_unification}. These mapping rules are applied as text filters. We use them to reformat the human-created gold transcripts for the ASR and NLP systems. For the annotation of \textit{ATCO2-test-set corpus}, we also defined an annotation manual that is reachable from the transcription platform \url{https://www.spokendata.com/atco2}.

\subsection{Lexicon}

The lexicon is a table that maps words into pronunciations (phoneme-strings). It is a resource used by the HMM-based ASR tools.
Our lexicon is based on the CMU Pronouncing Dictionary\footnote{Dictionary at: \url{http://www.speech.cs.cmu.edu/cgi-bin/cmudict}.}, which defines the phoneme set, and which is used as the training data for the grapheme-to-phoneme (G2P) module that synthesizes pronunciations of “new words”. We gather all possible words from the training corpora, and we add some other words from different resources. We synthesize the pronunciations by G2P model trained with the Phonetisaurus\footnote{More information in their GitHub repository: \url{https://github.com/AdolfVonKleist/Phonetisaurus}.} tool. 

The `spelled acronyms’ like “KLM” (pronounced as \textit{“k ey eh l eh m”}) are treated separately and represented as a single token (e.g.,`k\_l\_m’). We also add manually created pronunciations for some non-English words that cannot be guessed by the G2P model. All the `word tokens’ in the lexicon are in lower-case.
We keep only words relevant to ATC domain, i.e. words present in ATC transcripts or other resource. The lexicon contains 29k unique word-symbols. 

Our strategy to mitigate the out-of-vocabulary problem is based on enriching the lexicon as much as possible as part of the data preparation. We enriched the lexicon with a list of airline designators for callsigns (partly manually updated).\footnote{List taken from Wikipedia: \url{https://en.wikipedia.org/wiki/List_of_airline_codes}.} Also, we added all five-letter waypoint\footnote{A waypoint is an intermediate point or place on a route or line of travel, a stopping point or point at which an aircraft's course is changed.} names in Europe retrieved from open-source project Traffic\footnote{\textit{Traffic} project: \url{https://pypi.org/project/traffic/}.}. Finally, we introduced some additional words, such as countries, cities, airport names, airplane models and brands, ATC acronyms, etc.

\section{Data Collection}

This section describes the collection and pre-processing of the audio data and ATC metadata in the ATCO2 corpus. An overview of the data processing pipeline is given in Figure~\ref{fig:pipeline}.
First, the data is collected via very-high frequency (VHF) radio receivers that are owned by a community of volunteers. Then, the audio and metadata are uploaded to OpenSky Network (OSN) servers\footnote{OpenSky Network is a non-profit association based in Switzerland. It aims at improving the security, reliability, and efficiency of the airspace usage by providing open access of real-world air traffic control data to the public. The OpenSky Network consists of a multitude of sensors connected to the Internet by volunteers, industrial supporters, and academic/governmental organizations.}
via Internet. Finally, the collected data is processed on a ReplayWell server\footnote{The whole pipeline runs live in the following URL: \url{https://www.spokendata.com/atco2}.} via REST API, and part of it is selected for human annotation (see Appendix~\ref{appendix:transcription-pipeline}). The ReplayWell server hosts a major part of the data processing pipeline. We also rely on a community of volunteers for annotation.

\subsection{Data Feeders}
\label{subsec:feeders}

The data feeders are volunteers who capture ATC voice and upload it to OSN servers. The Data Feeders are typically aviation enthusiasts with possible prior operational experience, or people with an interest in aviation technologies (e.g., people doing domain related research, radio amateurs, etc.). 
To become a feeder, one needs to own a VHF receiver, which consists of an antenna, software defined radio (SDR) and a computer connected to the Internet. Affordable and popular options such as an RTL-SDR dongle and Raspberry Pi single board computer work sufficiently well in most cases.
Quality of recorded ATC data varies depending on the equipment utilized during its collection (properly tuned gain parameter, position of antenna, DSP processor in the radio receiver). As an example, an affordable setup can be built with a {\em Sirio Md 118-137} antenna and an {\em RTL-SDR} radio receiver dongle (RTL2832U with 8-bit analog-to-digital converter), this setup is similar to items in Figure~\ref{fig:atco-receiver}. For better quality, we recorded with a {\em Watson WBA-20} antenna and a {\em SDRPlay - RSP1A} radio receiver, which has a 14-bit analog-to-digital converter. 

In some countries, 
it might be prohibited by law to record air traffic management (ATM) related data. The data feeders should check the applicable regulations before recording and feeding the data to the Internet.
If you are interested in becoming a data feeder, please follow the instructions in the `feeder zone' website: \url{https://ui.atc.opensky-network.org/set-up}

\begin{figure}[t]
    \centering
    \includegraphics[width=0.5\linewidth]{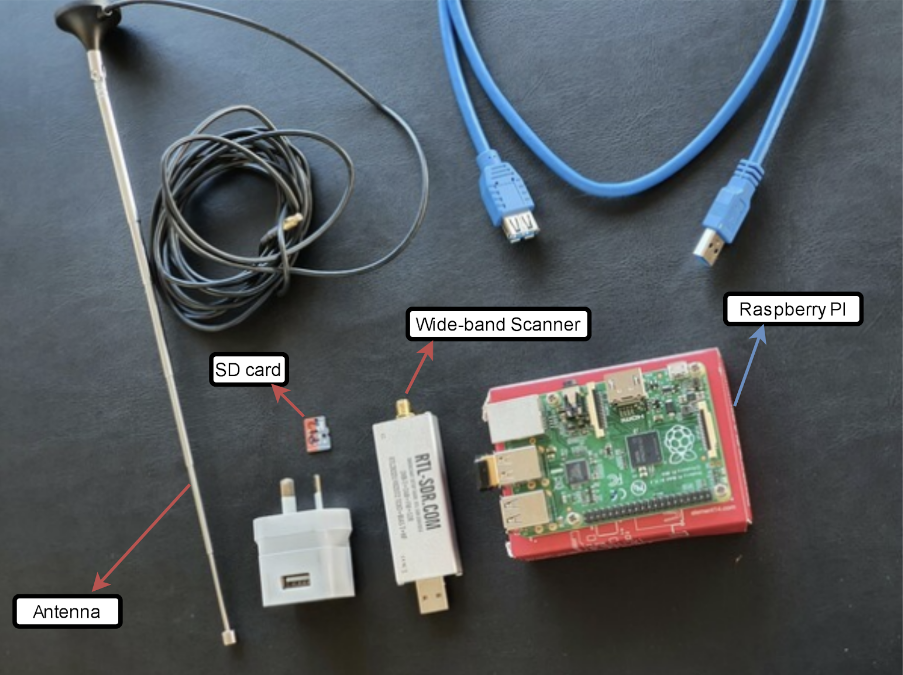}
    \caption{A set of items needed to set up a VHF receiver.}
    \label{fig:atco-receiver}
\end{figure}

\subsection{Data Annotators}
\label{subsec:transcriber}

The annotators are people who produce transcripts of ATC voice communications. These annotations also include assigning speakers roles and tagging of named entities. During the ATCO2 project, we relied on both the volunteers and paid transcribers. Volunteers with knowledge of ATC phraseology are ideal, but not strictly required.

Currently, we use our data processing pipeline (see Figure~\ref{fig:pipeline}), which generates the initial transcripts and NLP tags. Pre-transcribing with AI tools speeds-up the overall transcription process. If you are interested in becoming an annotator, please create an account in the SpokenData transcription platform: \url{http://www.spokendata.com/atco2}. All the transcribed data within ATCO2 project's life was packaged and released as the \textit{ATCO2-test-set corpus}. 
Both the data feeders and annotators will have access to the data they provided.\footnote{More information on the official Opensky Network website: \url{https://opensky-network.org}.}

\subsection{Data Processing Pipeline}
\label{subsec:data-processing-pipeline}

The steps from our automatic {\em data processing pipeline} in Figure~\ref{fig:pipeline} are briefly described here:

\textbf{Segmentation and demodulation:}
the RTL-SDR radio receiver tuned to a frequency provides a data stream in IQ format. The RTL-SDR software\,\footnote{RTL-SDR radio receiver software: \url{https://github.com/szpajder/RTLSDR-Airband.git}} has an in-built Voice Activity Detection (VAD) segmentation features. This is based on detecting abrupt changes of energy in the signal. The IQ signal is demodulated into a wave file with {\em csdr} software.\,\footnote{CSDR sofware defined radio: \url{https://github.com/ha7ilm/csdr}} The {\em csdr} software is configured to remove DC offset, and we don't use automatic gain control.
It is important to tune the {\em gain} parameter in the RTL-SDR software, so the audio is both well audible and not overboosted.

\textbf{Segment-based gain control:}
the signal from distant airplanes can be weak. We noticed speaker turns are often separated by spikes in the waveform. The spikes arise from the window-based DC-offset removal in {\em csdr}. We detect these spikes and increase volume in segments separated by the spikes when needed.

\textbf{Signal-to-noise ratio filtering:} 
the next processing step is “signal-to-noise ratio filtering”. We discard recordings that are too noisy, but audio files with moderate level of noise are not discarded.
We use WADA-SNR (Waveform Amplitude Distribution Analysis)~\cite{kim2008robust} to estimate the SNR. WADA-SNR is based on analysis of shape of distribution over samples in a speech waveform.
The non-speech parts are removed by a speech activity detection (SAD) tool~\cite{plchot2018analysis} with a `tight’ preset, leaving almost no non-speech parts marked as speech.

\textbf{Acoustic-based speaker diarization}: 
a single recording can have multiple speakers in it, so the per-speaker segments are identified by diarization. We do it before the automatic transcripts are generated, so the NLU modules always process segments of a single speaker. Also, the annotators are asked to eventually rectify the per-speaker segments.

For details of the acoustic diarization VBx model, the reader is referred to~\cite{landini2022bayesian}. This model uses a Bayesian hidden Markov model (BHMM) to find speaker clusters in a sequence of x-vectors. The x-vector extractor uses DNN architecture based on {\tt ResNet101}~\cite{landini2022bayesian}. In the first step, Agglomerative Hierarchical Clustering (AHC) is applied to the extracted x-vectors. Then, Variational Bayes HMM over x-vectors is applied using the AHC output.

\textbf{Automatic speech recognition:} 
our ASR system has been trained on several publicly available databases~\cite{UWB_ATCC,LDC_ATCC,PILSEN_ATC,ATCOSIM} and some private databases (AIRBUS, MALORCA). It is a hybrid ASR system trained with Kaldi. The system is covered in more details in Section~\ref{sec:asr}, and also in~\cite{kocour2021automatic}. The ASR output is confusion network, it is a `sausage-like' structure with lists of alternate words in bins, and word-confidences in each bin sum up to one.


\textbf{English language detection:} we deployed an {\em English language detection} system (ELD) to separate non-English utterances from the input stream of data. Specifically, we used an NLP-based 
system that processes ASR output transcripts with word confidences. This system was more robust and better than standard acoustic-based ELD system~\cite{szoke21_interspeech}.
Another benefit from using an NLP system is that it can jointly use logits or probabilities outputs from different ASR systems, which further can boost the results. Our ELD tool was previously covered in~\cite{kocour2021automatic, szoke21_interspeech}.

\textbf{Post-processing by NLP:} in ATCO2 project, we focused on extracting knowledge from the text produced by the ASR system. \textit{ATCO2-test-set corpus} contains rich metadata extracted with different NLP and NLU based modules. Specifically, we performed three tasks:

\begin{itemize}
    \item {\em Callsign recognition}: locate the callsign and convert it to code such as \verb|"KLM91G"|
    \item {\em ATCO/pilot classification}: decide who is speaking in the entire utterance
    \item {\em ATC-Entity recognition}: highlight the callsign, command and value entities in text
\end{itemize}

Further details about our NLP/NLU modules are covered in~\cite{kocour2021automatic}. Information about integration and pre-processing pipeline is in Appendix~\ref{appendix:full-pipeline}.

{\bf Dataflow statistics:} the statistics for our {\em data processing pipeline} are accessible in \url{https://www.spokendata.com/atco2}. The daily numbers summarize the amount of recordings entering the pipeline, being rejected for various reasons (e.g., too low SNR, non-English language detected), being automatically processed, or selected for human annotation.


\begin{table}[t]
  \caption{Train and test sets configuration for baseline experiments.
  $^{\dagger}$entire \textit{ATCO2-PL corpus} used for training our ASR modules, see Table~\ref{tab:databases}. $^\mathparagraph$this subset filters out recordings that do not contain speaker role tags (used for speaker role detection). However, we report results on the full \textit{ATCO2-test-set corpus} for the ASR experiments.}
  \label{tab:databases_experiments}
  \centering
  \begin{tabular}{ l c c c c }
    \toprule
    & \multicolumn{4}{c}{\textbf{ \cellcolor{Gray} Statistics}} \\
    \textbf{Dataset} & \multicolumn{1}{c}{Nb. samples\,[k]} & \multicolumn{1}{c}{Duration\,[h]} & \multicolumn{1}{c}{SNR\,[dB]} & \multicolumn{1}{c}{Public} \\
    \midrule
    {\em ATCO2-PL-set \textit{(train)}}$^{\dagger}$ & 3072 & 5281 & any & \checkmark \\
    {\em ATCO2-test-set \textit{(test)}}$^\mathparagraph$ & 3 & 3.4 & $\leq$15 & \checkmark \\
    \bottomrule
  \end{tabular}
\end{table}

\section{Datasets}
\label{sec:results}

Here, we describe in details the datasets for our baseline experiments. We evaluate on the {\em ATCO2-test-set corpus} as an in-domain test set, and {\em MALORCA-Vienna test set} as an unseen airport. The baseline systems are trained purely with the {\em ATCO2-PL-set corpus} and its automatic transcripts (i.e. pseudo-labels).




\subsection{ATCO2 databases}

\textbf{ATCO2-test-set corpus:} this dataset was built for development and evaluation of ASR and NLP technologies for English ATC communications. 
The dataset consists of English coming from LKTB, LKPR, LZIB, LSGS, LSZH, LSZB and YSSY airports.
We provide two partitions of the data, the \textit{ATCO2-test-set-1h corpus} and the \textit{ATCO2-test-set corpus}. The first corpus contains 1.1 hours of open-sourced transcribed annotations, and it can be accessed for free in \url{https://www.atco2.org/data}. The latter adds around 3 hours of annotated data and the full corpus will be available for purchase through ELDA in \url{http://catalog.elra.info/en-us/repository/browse/ELRA-S0484}. 
The amounts of data per airport are summarized in Table \ref{tab:stats-test-sets}.
The recordings of both corpora are mono-channel sampled at 16kHz and 16-bit PCM.
An example of the XML format for transcripts and tags is in Appendix\,\ref{appendix:sample-xml-tagged}.


\begin{table}[t]
    \caption{Stats about the collected databases per airport. Duration, SNR, language scores and contextual data columns report the mean and standard deviation (mean/\textbf{std}) per sample. Each recording/sample contains one or more segments (we provide timing information in RTTM format). $^\dagger$abbreviation in IETF format. $^{\dagger\dagger}$total number of segments and accumulated duration of speech (after voice activity detection) per airport.}
    \label{tab:stats-databases}
    \centering
    \resizebox{0.99\textwidth}{!}{
    \begin{tabular}{llllllllll}
        \toprule
        \multicolumn{2}{c}{\cellcolor{Gray} \textbf{Database}} & & \multicolumn{4}{c}{\cellcolor{Gray} \textbf{Metadata}} && \multicolumn{2}{c}{\cellcolor{Gray} \textbf{Contextual data}} \\
        \cline{1-2} \cline{4-7} \cline{9-10}
        \rule{0pt}{2.5ex} ICAO - Airport & Accent$^\dagger$ & & \# Segments$^{\dagger\dagger}$ & Dur. [sec] & SNR [dB] & Lang Score & & \# n-grams & \# entities \\
        \midrule
        \rowcolor{Gray}
        \multicolumn{10}{c}{\textbf{\textit{English Data (language score $\geq 0.5$) }}} \\
        \midrule
        
        EETN - Tallinn & et & & 79\,k{\small \textbf{/131\,hr}} & 6.0{\small \textbf{/3.4}} & 4.6{\small \textbf{/7.8}} & 0.96{\small \textbf{/0.08}} & & 104{\small \textbf{/26}} & 37{\small \textbf{/9}} \\
        EPLB - Lublin & pl & & <1\,k{\small \textbf{/<1\,hr}} & 13.3{\small \textbf{/8.0}} & 2.5{\small \textbf{/8.2}} & 0.94{\small \textbf{/0.11}} & & 19{\small \textbf{/10}} & 4{\small \textbf{/2}} \\
        LKPR - Prague & cs & & 999\,k{\small \textbf{/1762\,hr}} & 6.4{\small \textbf{/4.3}} & 14.2{\small \textbf{/8.2}} & 0.95{\small \textbf{/0.09}} & & 230{\small \textbf{/95}} & 70{\small \textbf{/30}} \\
        LKTB - Brno & cs & & 401\,k{\small \textbf{/888\,hr}} & 8.0{\small \textbf{/14.4}} & 4.1{\small \textbf{/15.7}} & 0.88{\small \textbf{/0.15}} & & 49{\small \textbf{/35}} & 15{\small \textbf{/10}} \\
        LSGS - Sion & fr-ch & & 168\,k{\small \textbf{/330\,hr}} & 7.1{\small \textbf{/4.8}} & 10.0{\small \textbf{/8.0}} & 0.87{\small \textbf{/0.15}} & & 56{\small \textbf{/23}} & 20{\small \textbf{/8}} \\
        LSZB - Bern & gsw & & 324\,k{\small \textbf{/699\,hr}} & 7.8{\small \textbf{/5.0}} & 15.4{\small \textbf{/10.7}} & 0.9{\small \textbf{/0.13}} & & 101{\small \textbf{/42}} & 36{\small \textbf{/15}} \\
        LSZH - Zurich & gsw & & 470\,k{\small \textbf{/921\,hr}} & 7.0{\small \textbf{/4.6}} & 7.8{\small \textbf{/7.7}} & 0.94{\small \textbf{/0.1}} & & 526{\small \textbf{/179}} & 169{\small \textbf{/55}} \\
        LZIB - Bratislava & sk & & 9\,k{\small \textbf{/24\,hr}} & 8.8{\small \textbf{/6.9}} & 5.4{\small \textbf{/8.7}} & 0.86{\small \textbf{/0.15}} & & 68{\small \textbf{/27}} & 22{\small \textbf{/8}} \\
        YBBN - Brisbane & en-au & & 105\,k{\small \textbf{/170\,hr}} & 5.8{\small \textbf{/4.1}} & 10.2{\small \textbf{/5.8}} & 0.93{\small \textbf{/0.1}} & & 268{\small \textbf{/86}} & 95{\small \textbf{/30}} \\
        YSSY - Sydney & en-au & & 49\,k{\small \textbf{/77\,hr}} & 5.7{\small \textbf{/9.2}} & 3.1{\small \textbf{/7.0}} & 0.92{\small \textbf{/0.11}} & & 495{\small \textbf{/148}} & 174{\small \textbf{/52}} \\
        others - others & others & & <1\,k{\small \textbf{/<1\,hr}} & 5.0{\small \textbf{/6.7}} & 4.0{\small \textbf{/7.4}} & 0.92{\small \textbf{/0.11}} & & 55{\small \textbf{/260}} & 16{\small \textbf{/78}} \\

        \midrule
        \rowcolor{Gray}
        \multicolumn{10}{c}{\textbf{\textit{Non-English Data (language score $< 0.5$) }}} \\
        \midrule
        
        EETN - Tallinn & et & & 2\,k{\small \textbf{/2\,hr}} & 4.0{\small \textbf{/2.4}} & 2.9{\small \textbf{/10.8}} & 0.3{\small \textbf{/0.14}} & & 95{\small \textbf{/30}} & 33{\small \textbf{/11}} \\
        EPLB - Lublin & pl & & <1\,k{\small \textbf{/<1\,hr}} & 13.1{\small \textbf{/2.7}} & -8.4{\small \textbf{/12.8}} & 0.2{\small \textbf{/0.13}} & & 17{\small \textbf{/7}} & 4{\small \textbf{/1}} \\
        LKPR - Prague & cs & & 105\,k{\small \textbf{/187\,hr}} & 6.4{\small \textbf{/5.4}} & 13.8{\small \textbf{/9.3}} & 0.18{\small \textbf{/0.16}} & & 217{\small \textbf{/97}} & 67{\small \textbf{/30}} \\
        LKTB - Brno & cs & & 214\,k{\small \textbf{/611\,hr}} & 10.3{\small \textbf{/19.2}} & 6.5{\small \textbf{/11.8}} & 0.15{\small \textbf{/0.15}} & & 56{\small \textbf{/33}} & 18{\small \textbf{/10}} \\
        LSGS - Sion & fr-ch & & 57\,k{\small \textbf{/83\,hr}} & 5.3{\small \textbf{/3.6}} & 9.8{\small \textbf{/9.3}} & 0.27{\small \textbf{/0.14}} & & 56{\small \textbf{/25}} & 20{\small \textbf{/8}} \\
        LSZB - Bern & gsw & & 42\,k{\small \textbf{/55\,hr}} & 4.7{\small \textbf{/3.4}} & 13.6{\small \textbf{/13.9}} & 0.3{\small \textbf{/0.13}} & & 102{\small \textbf{/45}} & 37{\small \textbf{/16}} \\
        LSZH - Zurich & gsw & & 36\,k{\small \textbf{/49\,hr}} & 5.0{\small \textbf{/4.1}} & 2.0{\small \textbf{/12.7}} & 0.25{\small \textbf{/0.15}} & & 485{\small \textbf{/180}} & 157{\small \textbf{/56}} \\
        LZIB - Bratislava & sk & & 10\,k{\small \textbf{/26\,hr}} & 9.0{\small \textbf{/7.6}} & 7.1{\small \textbf{/7.0}} & 0.18{\small \textbf{/0.15}} & & 72{\small \textbf{/26}} & 23{\small \textbf{/8}} \\
        YBBN - Brisbane & en-au & & 7\,k{\small \textbf{/10\,hr}} & 4.9{\small \textbf{/4.8}} & 5.9{\small \textbf{/12.3}} & 0.24{\small \textbf{/0.16}} & & 268{\small \textbf{/79}} & 95{\small \textbf{/28}} \\
        YSSY - Sydney & en-au & & 3\,k{\small \textbf{/3\,hr}} & 3.9{\small \textbf{/2.3}} & 2.7{\small \textbf{/10.5}} & 0.33{\small \textbf{/0.13}} & & 481{\small \textbf{/151}} & 169{\small \textbf{/53}} \\
        others - others & others & & <1\,k{\small \textbf{/<1\,hr}} & 5.2{\small \textbf{/6.4}} & 3.7{\small \textbf{/9.0}} & 0.26{\small \textbf{/0.15}} & & 0{\small \textbf{/0}} & 0{\small \textbf{/0}} \\

        \bottomrule
    \end{tabular}
    }
\end{table}

\begin{table}[h]
    \caption{\textit{ATCO2-test-set corpora} split by airports.}
    \label{tab:stats-test-sets}
    \centering
    \resizebox{0.60\textwidth}{!}{
    \begin{tabular}{lrrrr}
        \toprule\rowcolor{Gray}
              & \multicolumn{2}{c}{\cellcolor{Gray} {\it ATCO2-test-set}} 
                      & \multicolumn{2}{c}{\cellcolor{Gray} {\it ATCO2-test-set-1h}} \\
        \rowcolor{Gray}  
        {\bf ICAO - Airport} & {\bf sentences} & {\bf words} & {\bf sentences} & {\bf words} \\
        \midrule
        LKPR Prague   &  207 &  2686 & 102 & 1254 \\
        LKTB Brno     &   60 &   854 &  32 &  451 \\
        LSGS Sion     &  932 & 10183 & 256 & 2684 \\
        LSZB Bern     &  452 &  5908 & 172 & 2323 \\
        LSZH Zurich   &  640 &  8123 & 126 & 1764 \\
        LZIB Stefanik &  165 &  2256 &  79 & 1051 \\
        YSSY Sydney   & 1065 & 10434 & 102 & 1058 \\ 
        \midrule
        Sum           & 3521 & 40444 & 689 & 10585 \\
        
        \bottomrule
    \end{tabular}
    }
\end{table}

\begin{figure}
    \begin{subfigure}{.5\textwidth}
        \centering
        \includegraphics[width=.95\linewidth]{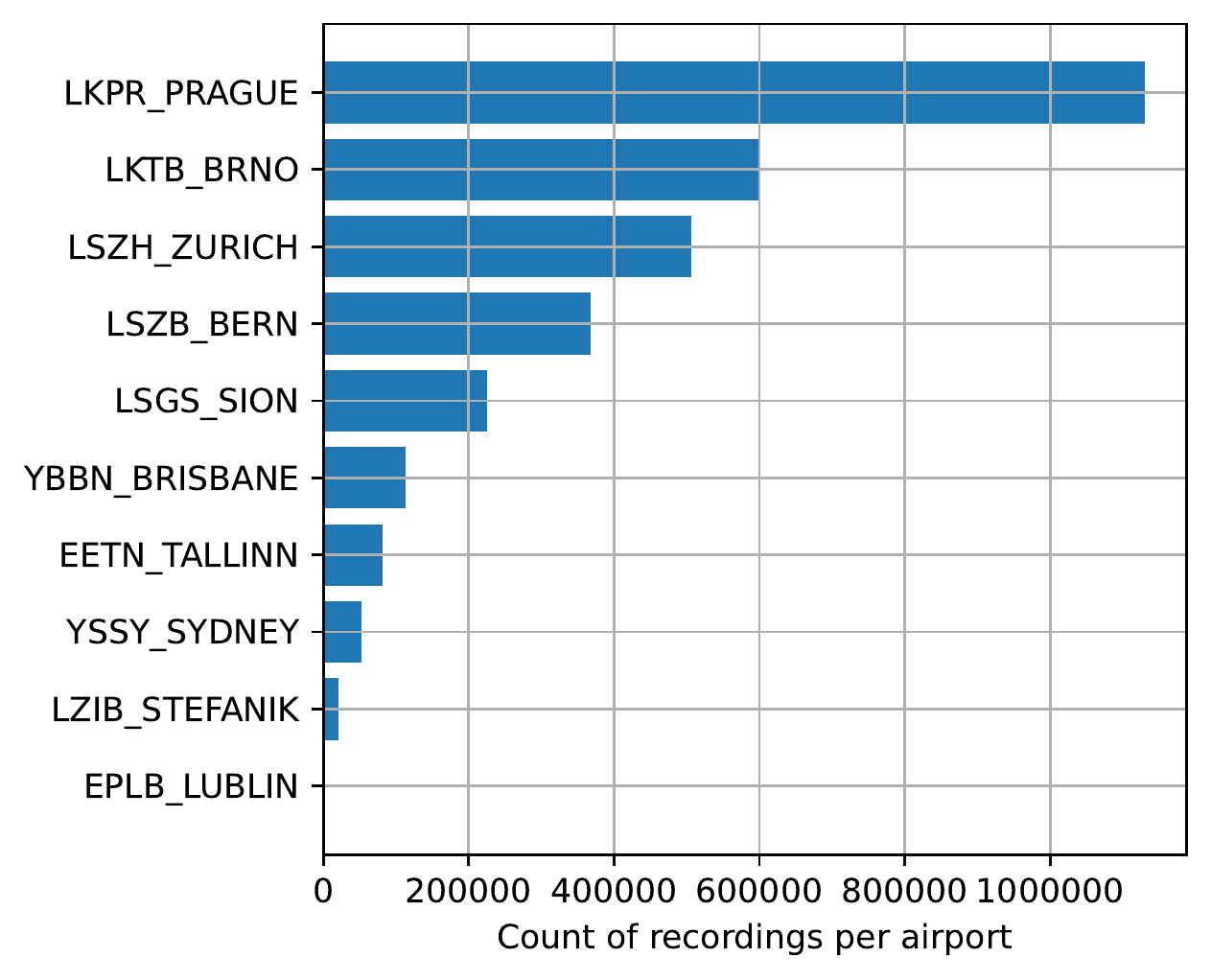}
        \caption{Number of recordings per airport}
        \label{fig:data_distribution_counts}
    \end{subfigure}%
    \begin{subfigure}{.5\textwidth}
        \centering
        \includegraphics[width=.95\linewidth]{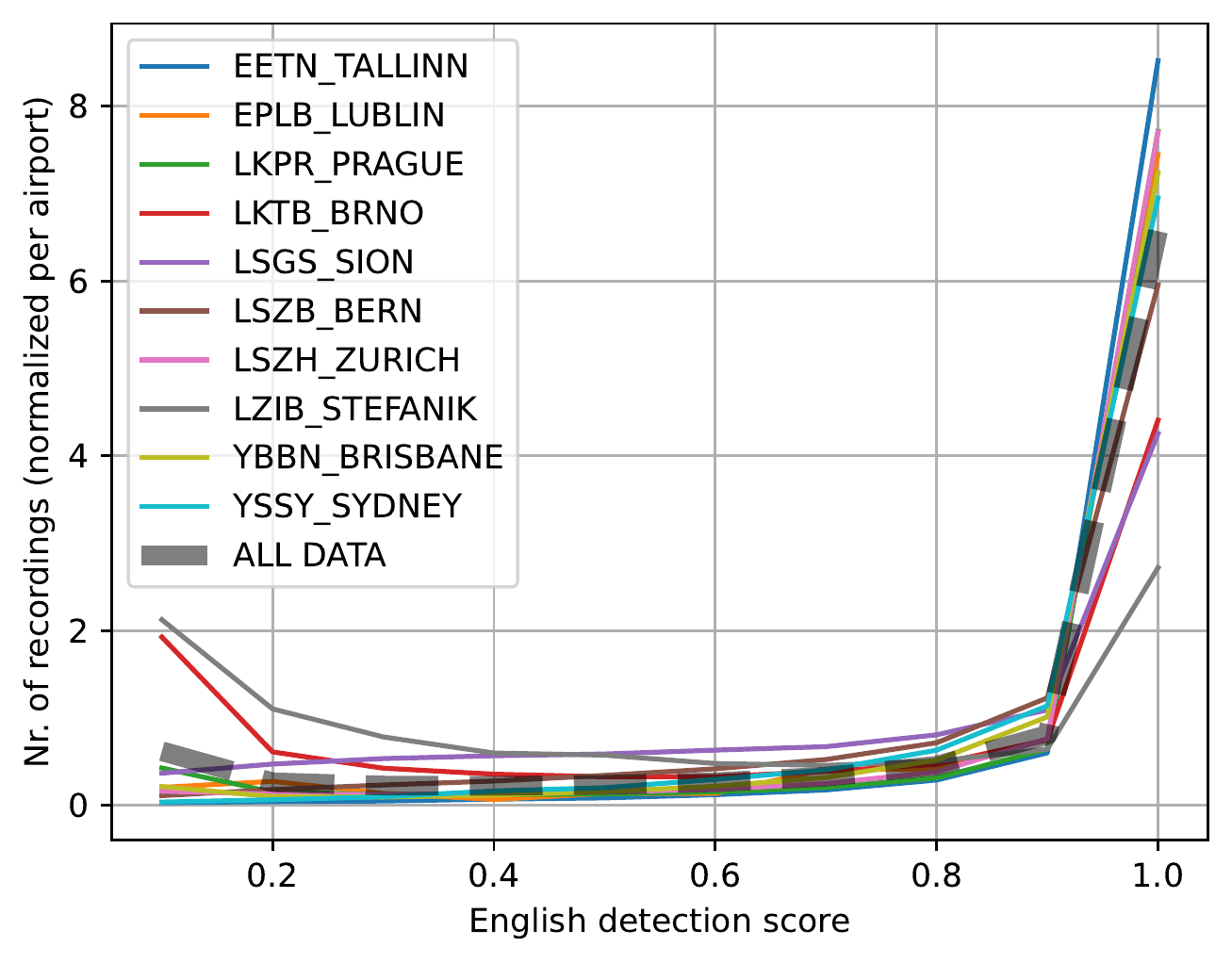}
        \caption{Distribution of English detection scores per airport}
        \label{fig:data_distribution_eld}
    \end{subfigure}
    \begin{subfigure}{.5\textwidth}
        \centering
        \includegraphics[width=.95\linewidth]{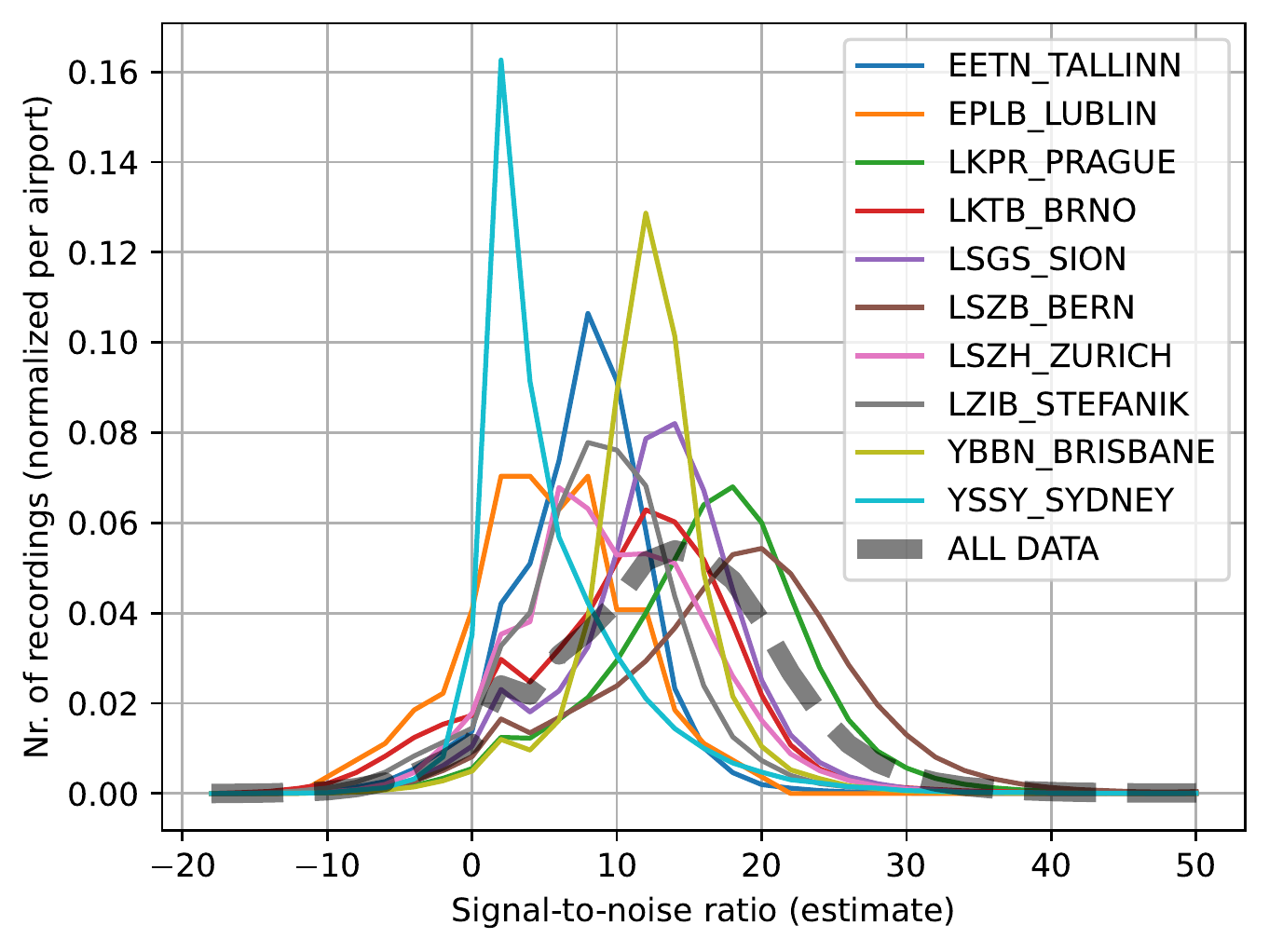}
        \caption{Distribution of Signal-to-noise ratio per airport}
        \label{fig:data_distribution_snr}
    \end{subfigure}%
    \begin{subfigure}{.5\textwidth}
        \centering
        \includegraphics[width=.95\linewidth]{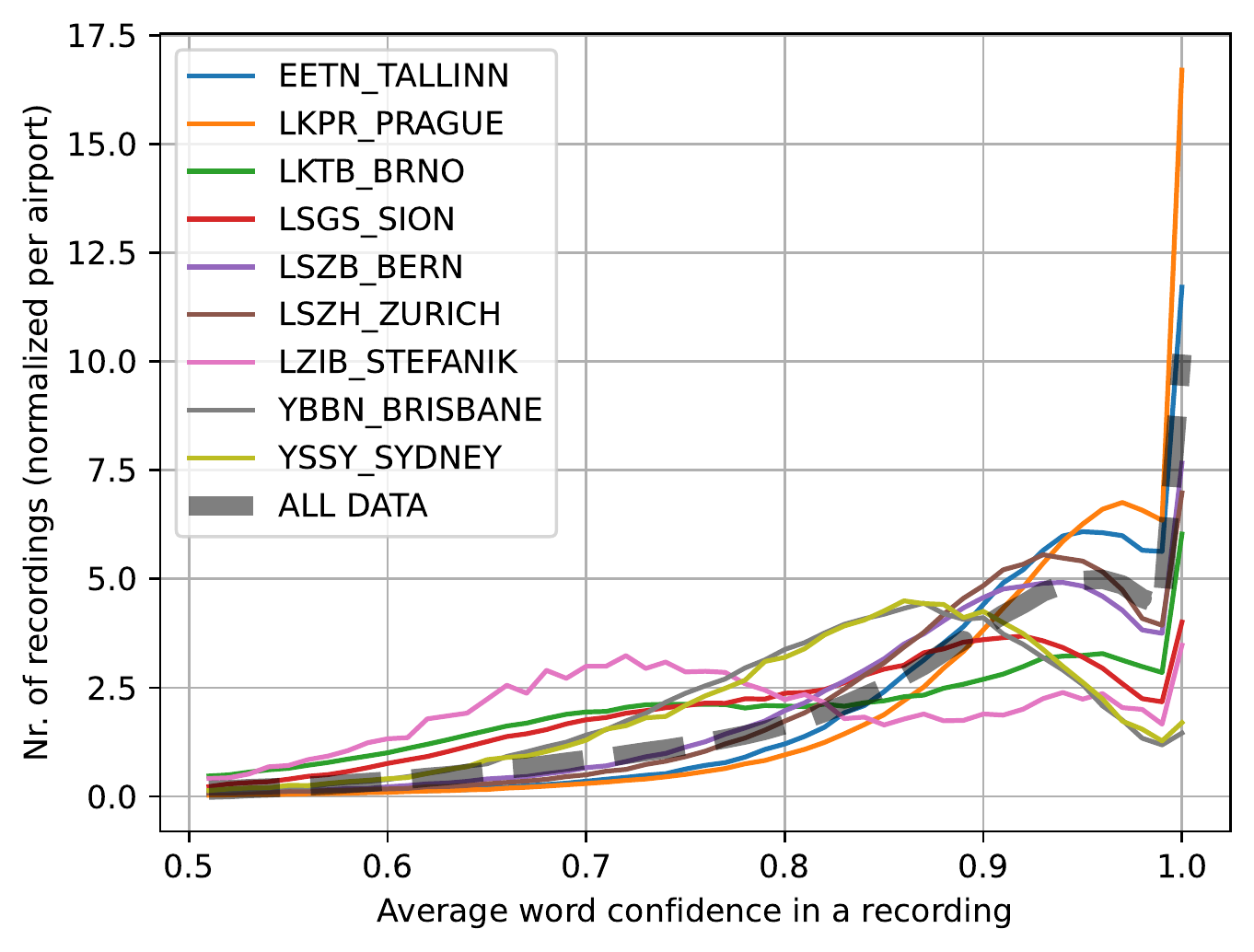}
        \caption{Distribution of Confidence scores per airport}
        \label{fig:data_distribution_confidence}
    \end{subfigure}
    \caption{Distribution plots of metadata for \textit{ATCO2-PL set corpus}.}
    \label{fig:data_distribution_graphs}
\end{figure}

\textbf{ATCO2-PL-set corpus:} ATCO2 project recorded a large database of ATC voice communications. Altogether, we collected over 5281 hours of ATC speech from different airports around the world (see Table~\ref{tab:stats-databases}). In total, we cover ten airports. 
Table~\ref{tab:stats-databases} depicts all this metadata per airport, further split by English language score. To the best of the author's knowledge, this is the largest and richest\footnote{By richest, we mean quality of annotations and amount of metadata per sample. Also, this is the first public database in the area of ATC that targets NLU tasks.} dataset in the area of ATC ever created that is accessible to the public. The automatic transcripts are stored as confusion network, stored as a {\em ctm} text format extended to have more words per line (confusion network bin):
\begin{verbatim}
<wav-id> <speaker> <t_begin> <dur> <word1> <conf1> <word2> <conf2> ...
LKPR_Tower_134_560MHz_20211223_154543 A 1.25 0.10 the 0.845 papa 0.042 ...
\end{verbatim}

Another view of the data is in Figure \ref{fig:data_distribution_graphs}, where we don't split by the English detection score. The distributions are from the full, 5281 hours dataset. In sub-figure \ref{fig:data_distribution_counts}, we see that the majority of data was recorded in Prague, Brno, Zurich, Bern, and Sion airports. This is because we started recording these airport's data early in the ATCO2 project. Next, we also have some data from Brisbane, Talinn, Sydney, Bratislava (STEFANIK) and Lublin. In~\ref{fig:data_distribution_eld}, we see that majority of our data has high English scores (the dashed line `ALL DATA'). There are some airports that have more non-English utterances: Brno, Bratislava (STEFANIK). And there are some airports, for which the distribution is more uniform than for others: Sion. We know for sure that local language is present at higher quantities in the data from Brno, Bratislava and Sion airports. And bigger airports usually have a policy to speak only in English, which explains low number of detections of non-English speech there. From \ref{fig:data_distribution_snr} we see that levels of noise differ across the airports. The cleanest signal is for Prague and Bern, while high levels of noise are from Sydney and Lublin, and some noise also for Tallinn and Bratislava. This indicates that the recording setup could be improved. And in \ref{fig:data_distribution_confidence} are the distributions of confidences of the automatic transcripts, assigned by the seed ASR system. Majority of the probability mass is in interval $(0.8, 1.0)$ (dashed curve `ALL DATA'). The highest confidence is assigned to Prague data (highest peak on right). The lowest confidence have the data from Bratislava, Brisbane and Sydney airports (distributions with leftmost modes). The higher tails with lower confidences are very likely caused by the non-English speech and noisy signal in the data.

\subsection{Private Databases}
\textbf{MALORCA Vienna test set:}
The MALORCA Vienna test set is used in baseline ASR experiments as an unseen airport. No Vienna data are in the {\em ATCO2-PL-set} that use for training the acoustic model and language model. On the other hand, MALORCA Vienna data were present in the training of the seed system for generating the automatic transcripts. So it both unseen and indirectly seen at the same time. The set consists of ATCO speech only, which normally has lower error rates than the  pilot speech~\cite{AIRBUS,pellegrini2018airbus}. 
The total amount of speech after VAD is 1.9 hours. The audio data is mono-channel sampled at 8kHz and 16-bit PCM.

\section{Automatic Speech Recognition}
\label{sec:asr}

The Automatic Speech Recognition (ASR) system has an audio signal as its input and produces text transcripts as its output.

At first, we trained a `seed ASR system' from several existing ATC databases. The seed ASR system is part of the `data processing pipeline' (see Section~\ref{subsec:data-processing-pipeline}). This ASR produced the automatic transcripts for the \textit{ATCO2-PL-set corpus} and also the initial transcripts for the \textit{ATCO2-test-set corpus} that were manually corrected.
During the ATCO2 project, we worked mainly with hybrid-based ASR systems, trained with the open-source toolkit Kaldi~\cite{povey2011kaldi}.

In hybrid speech recognizers, Hidden Markov Models support various speech rates, by allowing to "stay in a state" for some time via self-loop transitions. The acoustic scores come from a neural network, and the decoding process is based on searching in the matrix of acoustic scores for state-sequences with plausible transcriptions generated from a pre-compiled recognition network HCLG (i.e., a large HMM graph/model). Thus, HMMs provide a structure for mapping a temporal sequence of acoustic scores into a sequence of states \cite{morgan1993hybrid,bourlard1993connectionist}, from which the recognized words are extracted. 

\begin{figure}[t]
    \centering
    \includegraphics[width=\linewidth]{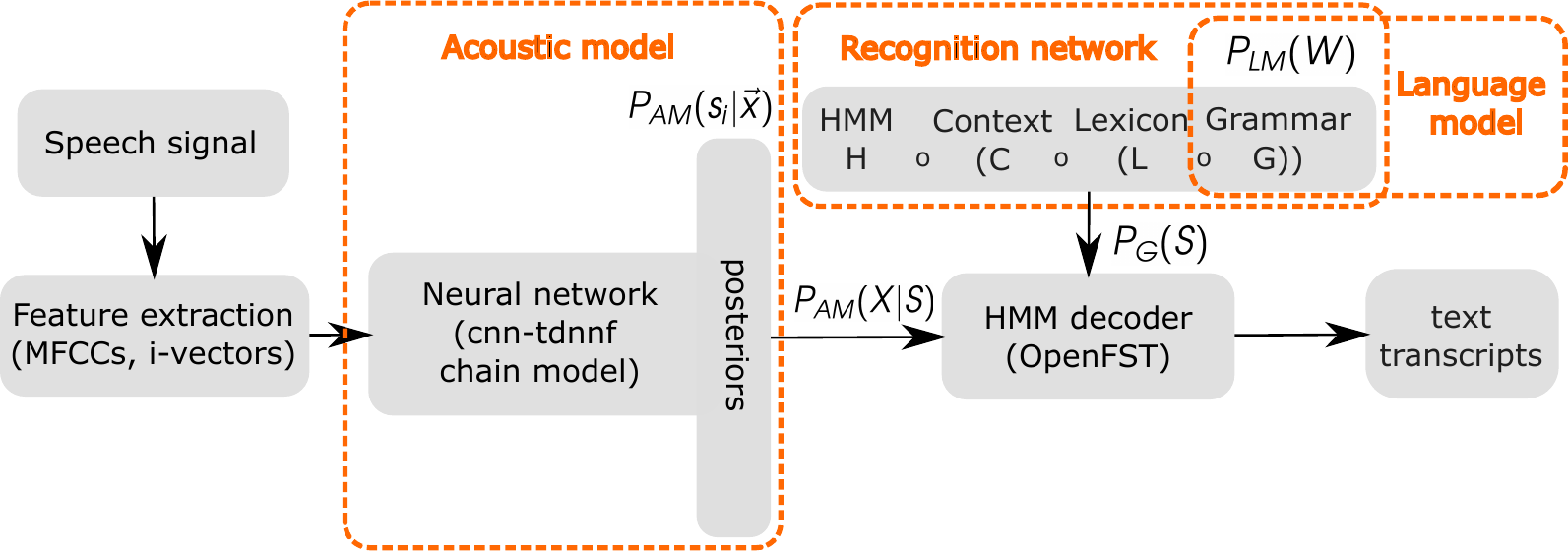}
    \caption{Hybrid-based ASR system. The inference pipeline consists of {\em feature extraction}, {\em acoustic matching} by acoustic model and {\em search} by HMM decoder that uses HCLG recognition network. On the output are text transcripts. Alternatively, the output can be a {\em lattice} (a graph with alternative decoded paths) or {\em confusion network} (time-sequence of bins with word-lists having scores).}
    \label{fig:hybrid_recognizer}
\end{figure}

Inference in a hybrid speech recognizer has three stages: {\bf feature extraction}, {\bf acoustic matching} and {\bf decoding}, the overall scheme is in Figure \ref{fig:hybrid_recognizer}. 

{\bf Feature extraction} compresses the waveform into a sequence of fixed-length vectors of low dimension, in our recipe we use high-resolution MFCCs with i-vectors~\cite{saon2013speaker} appended.

{\bf Matching of acoustic units} by acoustic model converts the input features into posterior probabilities of a closed set of acoustic units (phoneme states), whose time series forms the acoustic score matrix. In our recipe, we use `chain' model neural network (NN) trained by Lattice-free MMI~\cite{povey_lfmmi_IS2016}. The NN topology is  `cnn-tdnn' architecture with 6 {\tt conv-relu-batchnorm-layer} components followed by 9 {\tt tdnnf-layer} components~\cite{TDNN-F}, and 2 softmax layers with 2000 outputs each. The acoustic model consists of 12.9 million trainable parameters. 

{\bf Decoding} searches for the most likely word sequence $\hat{W}$ (transcription), in the matrix of acoustic scores.
The search explores HMM paths that exist in a recognition network, termed HCLG graph. The standard decoding algorithm is based on two ideas: {\em token passing} and {\em beam search}. The search combines scores from the acoustic model, language model and lexicon, as shown in equation \eqref{eq:decoding_formula}~:

\begin{equation}
    \hat{W} = \func{wrds}\left(\argmax_{S} P_{AM}(S|\V{X})^\kappa \; P_G(S)\right) \;. 
\label{eq:decoding_formula}
\end{equation}

\noindent
The acoustic model scores are the chain model posteriors $P_{AM}(S|\V{X})$, where $\V{X}$ is the time-series of input features and $S$ is an HMM state-sequence. The language model and lexicon scores are both represented in the graph score $P_G(S)$ that is present in the HCLG recognition network. $\kappa$ is an empirical scaling constant, for chain models the optimal $\kappa=1.0$. And the function $\func{wrds}(.)$ is reading word-sequence from the state sequence $S$ with the maximal score.

{\bf The HCLG graph} is a Weighted Finite State Transducer (WFST). The HCLG graph is composed of a language model graph G, pronunciation lexicon graph L, context dependency graph C and phoneme HMM graphs H. The HCLG graph contains graph costs $P_G$ that originate from its source graphs, while the most important source is the language model. This was the description of a hybrid ASR system. 

The other type of ASR systems are End2End systems. The End2End systems do not rely on HMMs and do not have a pronunciation lexicon. However, End2End systems require more training data to achieve good performance.
Hybrid systems remain one of the best and more flexible approaches for building ASR engines. The HMM-DNNs based ASR are used in the current state-of-the-art systems for ASR in ATC communications \cite{srinivasamurthy2017semi,kleinert2018semi,zuluaga2020automatic,zuluagagomez20_interspeech}. 

Hybrid-based ASR systems train independently the Acoustic model and the Language model. the language model is trained on a text corpus. This allows to incorporate text resources without the necessity to have the corresponding audio.
The hybrid ASR relies on a word-based lexicon, and words that are not in the lexicon or language model cannot be hypothesized by ASR decoder (Out-of-vocabulary word problem -- OOVs).

We use the same ASR system both for ATCOs and pilots. The training recipe and databases for our `seed ASR system' (including the train sets in Table~\ref{tab:databases}) are covered in~\cite{kocour2021automatic,zuluagagomez20_interspeech,zuluaga2020automatic,zuluagagomez21_interspeech}.
Briefly, we used AIRBUS, MALORCA Vienna, ATCOSIM, UWB-ATCC, LDC-ATCC, HIWIRE and N4-NATO databases.
In total, these form a database of $\approx$ 135 hours. We augmented this database with noises captured from LiveATC. And the data were further augmented by speed perturbation. Due to data license issues with some databases, this ASR system can be only used for research.

In a later stage of the ATCO2 project, we experimented with {\bf contextual adaptation} and {\bf semi-supervised training}. We later integrated these technologies into the `seed ASR system'. 

The {\bf contextual adaptation} improves the accuracy of the ASR system by feeding-in a rapidly changing contextual information. Based on surveillance data, we suggested a list of nearby callsigns into the recognizer~\cite{kocour21_interspeech}. This was done by applying a boosting WFST graph to HCLG or lattice. In HCLG boosting, we give score discounts to individual words, while in Lattice boosting, the score discounts are given to word sequences. The lattice boosting was used also when generating the automatic transcripts for {\em ATCO2-PL-set corpus}. Also, in~\cite{nigmatulina2022two,nigmatulina2021improving} lattice boosting and language model boosting are explored.

The {\bf semi-supervised training} was used to improve ASR accuracy by retraining the acoustic model~\cite{kocour2021automatic} on a mixture of manually and automatically transcribed data. ATCO2 data with automatic transcripts were mixed with transcribed data from other databases. We used per-frame gradient weighting by word confidences to de-weight data with unreliable transcripts. We further performed experiments in~\cite{zuluagagomez21_interspeech}. Here, we applied callsign boosting when generating automatic transcripts for semi-supervised learning, and we obtained 17.5\% relative WER improvement measured on the callsign words.

\subsection{Baseline experiments}

During the ATCO2 project, we collected 5281 hours of ATC audio data from several airports. We processed the data with our automatic pipeline (see Figure~\ref{fig:pipeline}) that filters the data and produces automatic transcripts.
Inside the pipeline, there is an ASR system that is described in Section~\ref{sec:asr} and also in our previous work~\cite{kocour2021automatic}.

The purpose of these baseline experiments is to demonstrate what can be achieved with the data we collected and released in ELDA catalogue.
From these automatic transcripts, we can bootstrap and build a new ASR system, without having licensing problems that exist for some other databases (Table \ref{tab:databases}). We built a new language model from all the generated transcripts. And, we experimented with training acoustic models on various subsets of the audio data. We re-used the lexicon
from the `seed ASR system'. 

The baseline experiments are described in Table~\ref{tab:kaldi_result_alldata}, where we computed Word Error Rate (WER) on three test sets: {\em ATCO2-test-set}, {\em ATCO2-test-set-1h} and {\em MALORCA Vienna} test set. Each model is tagged with the number from the first column of Table~\ref{tab:kaldi_result_alldata}.
The MALORCA Vienna test set represents clean ATCO speech from an unseen airport.

\textbf{Analysis of ASR systems from Table~\ref{tab:kaldi_result_alldata}: } In 1) we built the acoustic model and language model on all the data in the ELDA package, including the data that the English detector identified as non-English. In 2) we excluded the non-English data, and from now on, the Language Model is always trained from transcripts of this 4500 hours dataset (except for seed system). In 3) we set the filtering thresholds higher to >0.7 ELD (English detection), >0 dB SNR and >0.8 CNET score (average word confidence in a confusion network of recording). In WER results for 1) 2) 3), we see that the \textit{ATCO2 test sets} results stay similar, while the WER for MALORCA test set improves with stronger data filtering.
In 4) we realized that it is not a good idea to discard too much noisy data by filtering >16 SNR.
Next, in 5) we randomly selected 3600 hours from the 4500 hours dataset. This was to cross-check with the filtering we previously used in 3). To our surprise, the results are marginally better when randomly selecting the data. 
Next, in 6) 7) 8) we continued randomly selecting subsets from the 4500 hours dataset. This degraded the performance of MALORCA Vienna on 7) 8) by up to 2\% WER. From the results, we notice that WER for ATCO2 test-sets is nearly constant, except 4). It seems that WER in the automatic transcripts is an important factor. The automatic transcripts are used as training targets, and the WER in transcripts pre-determines the performance of the trained system. The amount of training data is possibly less important, however the generalization to a new airport (MALORCA Vienna) is better with larger volumes of training data of 2500 or 3600 hours in ASR systems 5) 6).
For completeness, we also add WER of the seed system 9). The seed system had few percent higher WER for ATCO2 test-sets. For MALORCA Vienna, the seed system works as good as 4.8\% WER, as MALORCA Vienna corpus was present in its training data.

\begin{table}[t]
    \caption{Performance on \textit{ATCO2-test-set corpus}. The ASR system is built from the automatic transcripts of the \textit{ATCO2-PL-set corpus}. We use two \textit{ATCO2-test-set corpus} splits, and MALORCA Vienna~\cite{kleinert2018semi,srinivasamurthy2017semi} as an unseen airport. The data filtering is done according to: ELD (English language detection), SNR (signal-to-noise) ratio, and CNET (average word confidence in the recording).}
    \label{tab:kaldi_result_alldata}
    \centering
    \resizebox{0.99\textwidth}{!}{
    \begin{tabular}{ c p{1.4cm} p{0.1cm} p{1.0cm} p{1.0cm} p{1.7cm} p{0.1cm} ccc p{4cm} }
        \toprule
        \rowcolor{Gray} &  &  & \multicolumn{3}{c}{\textbf{WER}} &  & \multicolumn{4}{c}{\textbf{Data Selection Method}} \\
        \cline{4-6} \cline{8-11}
        \rowcolor{Gray} \rule{0pt}{3ex} {\bf System} & {\bf Training hours} &  & {\footnotesize \bf ATCO2 test-set (4h)} & {\footnotesize \bf ATCO2 test-set 1h} & {\footnotesize \bf MALORCA Vienna} &  & \textbf{ELD} & \textbf{SNR} & \textbf{CNET} & \textbf{Note:} \\
        \midrule
        1) & 5281 &  & 22.3 & 15.8 & 11.1 &  & any & any & any & all ATCO2 data \\ 
        \midrule
        2) & 4500 &  & 22.5 & 15.7 & 10.0 &  & \textgreater 0.5 & any & any & remove non-English \\
        3) & 3600 &  & 22.5 & 15.8 & 9.3 &  & \textgreater 0.7 & \textgreater 0 & \textgreater 0.8 & remove low-confidence \\
        4) & 1500 &  & 23.4 & 16.7 & 11.9 &  & \textgreater 0.5 & \textgreater 16 & any & remove low SNR (\textless{}16) \\ 
        \midrule
        5) & 3600 &  & 22.4 & 15.4 & 9.0 &  & \textgreater 0.5 & any & any & random from 4500 hour set \\
        6) & 2500 &  & 22.6 & 15.8 & 9.0 &  & \textgreater 0.5 & any & any & random from 4500 hour set \\
        7) & 1500 &  & 22.5 & 15.8 & 10.6 &  & \textgreater 0.5 & any & any & random from 4500 hour set \\
        8) & 500 &  & 22.5 & 15.7 & 11.0 &  & \textgreater 0.5 & any & any & random from 4500 hour set \\
        \midrule
        9) & 135+700 &  & 26.6 & 18.6 & 4.8 &  & - & - & - & seed system \\
        \bottomrule
    \end{tabular}
    }
\end{table}

\section{Natural Language Processing and Understanding}
\label{sec:nlp_nlu}

Until the previous decade, research on ATC was directed at only transcribing as accurate as possible the dialogues between ATCOs and pilots. However, transcription is only one part of the story and further information, such as, entity highlighting (also known as intent and slot filling) or speaker role detection is imperative in real-life ATC control rooms. The process of parsing these high-level entities from ATC audio can be seen as SLU, or from text as NLU.

Previous work has already explained several NLP tasks in the area of ATC. For instance,~\cite{lin2021spoken} describes a set of entities and elements that are present in ATC communications that are of special interest, e.g., commands and instructions. The authors advise that a real-life system should be composed of an ASR module to obtain the word-level transcripts of the communication. Later, a subsequent system should extract ATC-related key entities and then parse them into a specific grammar. 
We redirect the reader to~\cite{helmke2018ontology}, which developed an ATC-structured grammar accepted by several European institutes. Furthermore, in~\cite{lin2021spoken}, the process of extracting key entities from audio is summarized in an entire pipeline composed of three submodules. Namely, speaker role detection, intent classification and, slot filling (analogous to NER but on audio level). They aim at inferring the near-future air traffic dynamics, which can aid ATCOs in their daily task. In addition, this system can notice communication errors caused by one of the speakers, also known as  hear or read back errors. Some exploratory work addressing NLP and NLU on the framework of HAAWAII and ATCO2 projects (see Table~\ref{tab:databases}) is described in~\cite{prasad2021grammar,zuluaga2021bertraffic}. 

In this section, we describe our baselines for two tasks related to NLP and NLU.\footnote{As we work on top of ASR transcripts, these tasks can be also cataloged as spoken language understanding.} In ATC, in addition to transcripts generated by an ASR system, we can also extract rich metadata from the transcripts and audio. Some examples are (but not limited):

\begin{itemize}
    \item \textcolor{blue}{\checkmark} What are the high-level entities in the communication? $\rightarrow$ named-entity recognition (NER) or slot filling (SF). Previous work in~\cite{nigmatulina2022two} and covered in Section~\ref{subsec:ner_system},
    \item \textcolor{blue}{\checkmark} Who is talking? ATCO or pilot $\rightarrow$ speaker role detection (SRD), sequence classification. Early work on~\cite{prasad2021grammar}, and covered in Section~\ref{subsec:spk_id},
    \item \textcolor{magenta}{\XSolidBrush} Is the pilot responding the correct information? $\rightarrow$ read-back error detection. Our previous work in~\cite{helmke2021readback},
    \item \textcolor{magenta}{\XSolidBrush} Is the communication being uttered in English language? $\rightarrow$ English language detection (ELD). Our previous work in~\cite{szoke21_interspeech}.
\end{itemize}

We present baselines only on the above items marked with \textcolor{blue}{\checkmark}, while the items marked with \textcolor{magenta}{\XSolidBrush} are, either, covered in previous work or are left as future research directions. Generally speaking, extracting the above-mentioned information could allow to further fulfill other ATC tasks, e.g., pre-filling radar labels in the ATC control room. Or, for example, decrease the workload of ATCOs and increase their efficiency by automating manual and effortful processes. In addition, reducing the overall probability of incidents and accidents due to air traffic management erroneous procedures is a supplementary by-product of introducing AI tools in the ATC control rooms.

\begin{figure}[t]
    \centering
    \includegraphics[width=0.95\linewidth]{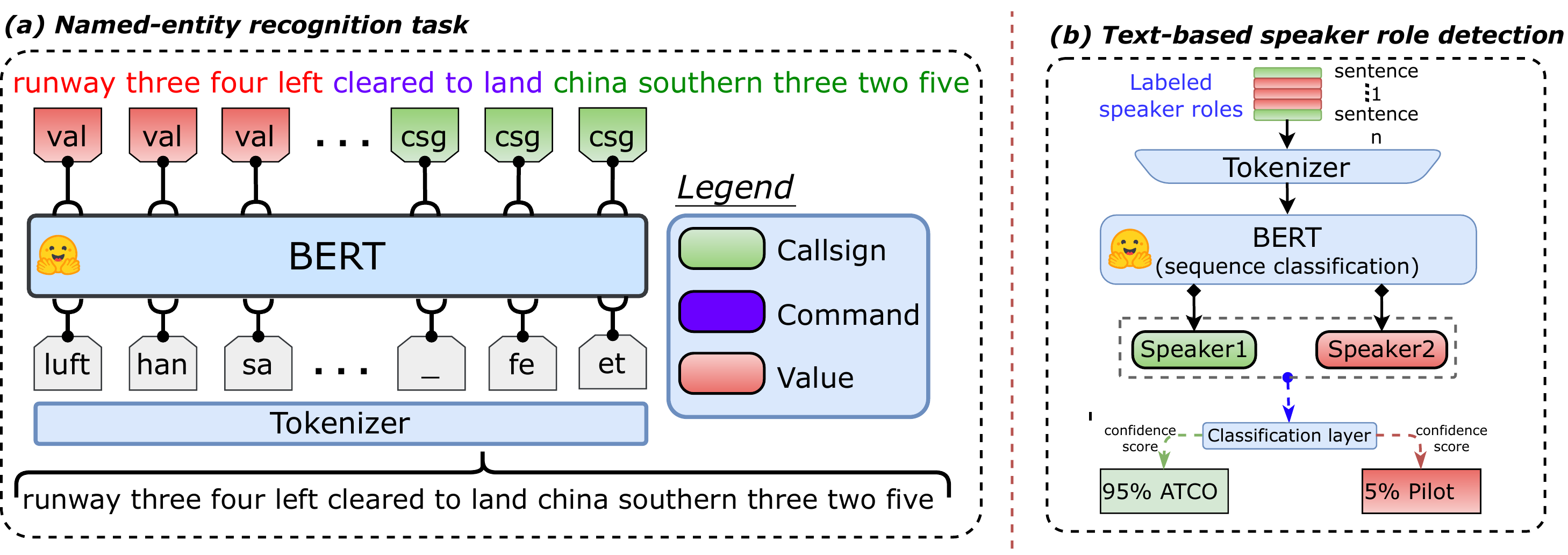}
    \caption{(a) Named entity recognition and (b) speaker role detection based on sequence classification (SC) for ATC utterances. Both systems are based on fine-tuning to ATC tasks from a pre-trained BERT~\cite{devlin2018bert} model. The NER systems recognizes callsign, command and values, while the SC assigns a speaker role to the input sequence.}
    \label{fig:ner-module}
\end{figure}

\subsection{Named Entity Recognition}
\label{subsec:ner_system}

Named entity recognition, or NER, is one of the most explored tasks in the field of information extraction and NLP~\cite{vajjala2022we}. NER aims to locate and classify entities in unstructured text into pre-defined classes or categories. Examples are, persons or organization names, expressions, or, for instance, callsigns or commands in ATC. Initially, NER was based on handcrafted lexicons, ontology, dictionaries, and rules~\cite{grishman1996message}. Even though these systems were interpretable and understandable, they were prone to human errors. Collobert et al.~\cite{collobert2011natural} introduced machine learning-based methods for text processing in topics such as part-of-speech tagging, chunking, NER, and semantic role labeling. Further interesting works on NER are \cite{piskorski2017first} focusing on multilingual NER for slavic languages, and \cite{yadav2018survey} presenting a broad survey of NER methods. 
In practice, a NER system can be crafted by fine-tuning a pre-trained LM, e.g., BERT~\cite{devlin2018bert}, RoBERTa~\cite{liu2019roberta}, or DeBERTa~\cite{he2021deberta}. Nonetheless, these models are data hungry and need expensive GPUs during its training and inference. Further work has been directed at reducing their computational footprint, by performing, for example, knowledge distillation~\cite{sanh2019distilbert}. 

Air traffic control communications typically carry structured information, including callsigns, commands and values. These can be seen as `named entities'. \textit{ATCO2-test-set corpus} provides annotation on the word level that assigns pieces of text to these specific classes. We developed a baseline system to extract such information from ASR utterances, as depicted in Figure~\ref{fig:ner-module}. An early implementation of this system was covered in~\cite{nigmatulina2022two}. However, these experiments were carried over private databases, so it is difficult to compare with our current results. That is why we base our experiments in~\cite{nigmatulina2022two}, but we go beyond by open sourcing scripts to fine-tune a NER model with \textit{ATCO2-test-set corpus}.

\subsubsection{Experimental Setup}
\label{subsec:ner_es}

Our experiments are carried out with \textit{ATCO2-test-set corpus} only, for both, training and evaluation.\footnote{We provide in the GitHub repository the utterance IDs splits utilized for these experiments.} The main reason is that none of the public databases from Table~\ref{tab:databases} contain NER annotations. As a workaround, we implemented a simple k-fold cross-validation scheme. We define $K=5$ folds, with a 70/10/20 ratio for train/dev/test subsets, respectively. We use ground truth ASR transcripts for training and testing NER.

First, we download a powerful pre-trained LM, BERT\footnote{We use the pre-trained version of \texttt{bert-base-uncased} with 110 million parameters for all the experiments.}~\cite{devlin2018bert}, from HuggingFace~\cite{wolf2020transformers,lhoest2021datasets}. 
We append a linear layer with a dimension of 8 (following the IOB format, i.e., two outputs for each NER class) on top of the last layer of the BERT model.
The model is later fine-tuned on the NER task, with each Fold $K$ of the train splits. 
Each model is fine-tuned on an NVIDIA GeForce RTX 3090 for 10k steps. During experimentation, we use the same learning rate of $\gamma=5\mathrm{e}{-5}$ with a linear learning rate scheduler. Dropout~\cite{srivastava2014dropout} is set to $dp=0.1$ for the attention and hidden layers, while Gaussian Error Linear Units (GELU) is used as activation function~\cite{hendrycks2016gaussian}. We also employ gradient norm clipping~\cite{pascanu2013difficulty}. We fine-tune each model with an effective batch size of 32 over 50 epochs with AdamW~\cite{loshchilov2018decoupled} optimizer ($\beta_1{=}0.9, \beta_2{=}0.999$, $\epsilon{=}1\mathrm{e}{-8}$). 

\subsubsection{Results}

We report the results obtained from the 5-fold cross validation experiments. We split the results by tags, namely, callsign, command and values. For each of them, we report precision, recall and F1-scores in Table~\ref{tab:ner_results}. 
We obtained an average of 0.97, 0.82 and 0.87 F1-score for callsign, commands and values. 
We observed that the command class was the most challenging among the three classes. 
We believe this is because commands contain extra complexity in comparison to callsigns and values. For example, in some cases the ATCOs or pilots use several commands, or these are sometimes mixed in the same utterance. In contrary, callsigns follow a standard form, composed of an airline designator, numbers, and letters (spelled in ICAO phraseology). Values are composed of cardinal numbers and some standard words, e.g., `flight level'. 
We also noted a significant irregularity in performance for the command class between the 5 folds (see column: Command in Table~\ref{tab:ner_results}). For example, worse $\rightarrow$ best scenario on F1-score was 0.79 $\rightarrow$ 0.85, almost a six-point drop. A five-point drop is also seen in precision and recall. These results are seen when comparing fold 2 (best) against fold 4 (worst).

In conclusion, the results from Table~\ref{tab:ner_results} are the first official baseline for NER\footnote{After extensive research, to authors' knowledge, this is the first official baseline on NER for air traffic control communications. We have not found any other work that is, both open-source and that targets NER.} on the \textit{ATCO2-test-set corpus}. However, there is room for improvement. For instance, implementing semi-supervised learning or data augmentation should bring robustness and yield higher performance. Similarly, one can pretrain the LM directly on ATC text rather than standard English text, which should bring in additional benefits. We leave this line of research for future work.

\begin{table}[t]
    \caption{Different performance metrics for callsign, command and values classes of the NER system. Metrics reported for each of the 5-fold cross-validation scheme on \textit{ATCO2-test-set corpus} with a \texttt{bert-base-uncased} model. @P, @R and @F1, refers to precision, recall and F1-score, respectively. Numbers in \textbf{bold} refers to the top performance per column among folds. $^\dagger$mean score over the 5 folds.}
    \label{tab:ner_results}
    \centering
    
    \begin{tabular}{c lllllllllllll}
        \toprule
        \multirow{2}{*}{\textbf{Fold}} &  & 
        \multicolumn{3}{c}{\cellcolor{Gray}\textbf{\textcolor{teal}{Callsign}}} &  & \multicolumn{3}{c}{\cellcolor{Gray}\textbf{\textcolor{purple}{Command}}} &  & \multicolumn{3}{c}{\cellcolor{Gray}\textbf{\textcolor{red}{Values}}} &  \\
        \cline{3-5} \cline{7-9} \cline{11-13}
         \rule{0pt}{3ex} &  & @P & @R & @F1 &  & @P & @R & @F1 &  & @P & @R & @F1 &  \\
        \midrule
        1 &  & 0.97 & 0.98 & 0.97 &  & 0.80 & 0.81 & 0.81 &  & 0.86 & 0.86 & 0.86 &  \\
        2 &  & 0.97 & 0.98 & 0.97 &  & \textbf{0.83} & \textbf{0.86} & \textbf{0.85} &  & 0.86 & 0.89 & 0.87 &  \\
        3 &  & 0.97 & 0.97 & 0.97 &  & 0.81 & 0.85 & 0.83 &  & \textbf{0.87} & 0.87 & 0.87 &  \\
        4 &  & \textbf{0.98} & \textbf{0.98} & 0.98 &  & 0.78 & 0.80 & 0.79 &  & 0.85 & \textbf{0.90} & 0.87 &  \\
        5 &  & 0.97 & 0.98 & \textbf{0.98} &  & 0.80 & 0.83 & 0.81 &  & \textbf{0.87} & 0.89 & \textbf{0.88} &  \\
        \midrule
        \rowcolor{Gray} AVG$^\dagger$ &  & 0.97 & 0.98 & 0.97 &  & 0.80 & 0.83 & 0.82 &  & 0.86 & 0.88 & 0.87 & \\
        \bottomrule
    \end{tabular}
\end{table}

\begin{wrapfigure}{R}{0.6\linewidth}
    \begin{center}
        \includegraphics[width=0.6\linewidth]{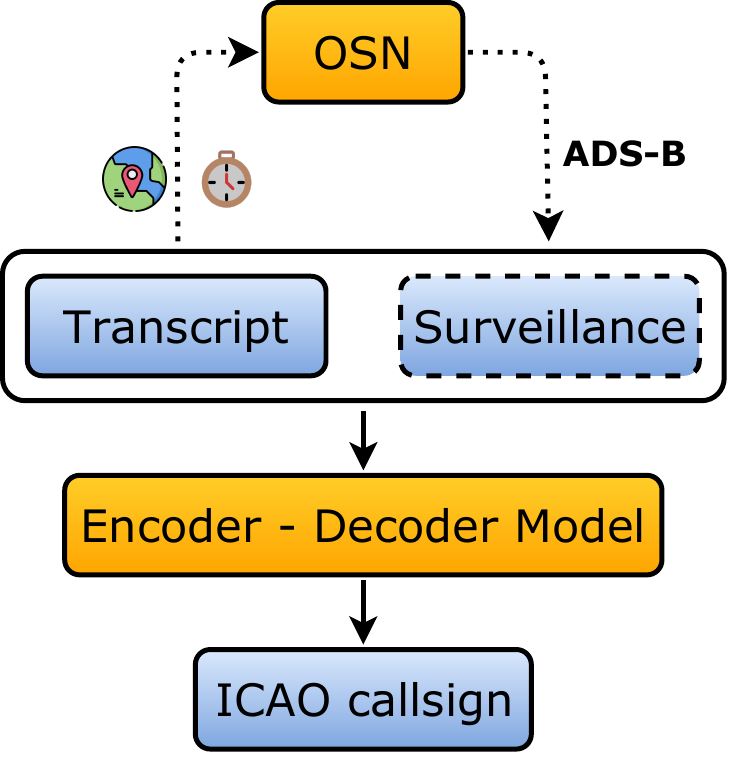}
        \caption{Proposed callsign recognition and understanding system. The dotted path marks the optional surveillance retrieval via OSN with the aid of the transcripts timestamp and VHF receiver location. Taken from~\cite{blatt2022callsigns}.}
        \label{fig:callsign-recognition}
    \end{center}
\end{wrapfigure}

\subsection{Callsign Recognition and Understanding}
\label{subsec:callsign_extraction}

The Named Entity Recognition from previous section is capable to select words which form a callsign (i.e., highlight `swiss two six eight nine'). However, {\em ICAO Callsign Extraction} produces callsign directly in ICAO format (e.g., SWR2689), which is more useful for applications. This is not trivial because callsigns get commonly shortened, if the situation is obvious (e.g., `swiss two six eight nine' $\rightarrow$ `six eight nine', or `swiss eight nine'). And the underlying ASR produces errors in its automatic transcripts.

In the project, we explored two approaches. In \cite{blatt2022callsigns}, the ICAO callsign is retrieved by a BERT-based Encoder-Decoder neural network. This system directly takes outputs from an in-domain ASR system and extracts the ICAO callsign without relying on Named Entity Recognition as an intermediate step. The model uses a list of callsigns from surveillance data as context information, and it can return an ICAO callsign that is not present in the list. The overall approach is depicted in Figure~\ref{fig:callsign-recognition}.

The second approach \cite{nigmatulina2022two} performs NER to extract the callsign within the sentence, which is later ranked by Levenshtein distance with the ones in the callsign list from the surveillance data. This approach always selects a callsign from the list. We showed that boosting callsigns with the combination of ASR and NLP methods eventually leads up to 53.7\% of an absolute, or 60.4\% of a relative, improvement in callsign recognition.

\subsection{Speaker Role Detection}
\label{subsec:spk_id}

In NLP, text classification or sequence classification (SC) is a task that assigns a label or a class to a sequence of words~\cite{he2020survey,zhou2015pattern}. The hypothesis is that the words within the given text share a common role and meaning inside the sentence's grammatical structure. One of the most acknowledged forms of SC is sentiment analysis, which assigns a label like positive, negative, or neutral to a sequence of text embeddings\footnote{However, a sequence of acoustic embeddings can also be used for SC, e.g., emotion classification in raw speech~\cite{purohit2022comparing}.} \cite{birjali2021comprehensive}.
Nowadays, state-of-the-art SC systems are based on the well-known Transformer, e.g., BERT~\cite{devlin2018bert} or RoBERTa~\cite{liu2019roberta}. Akin to NER, SC is considered a downstream task operating on ASR output. 

In ATC, the dialogues are built on top of a well-defined lexicon and dictionary, which follows a simple grammar. This standard phraseology has been defined by the ICAO~\cite{allclear} to guarantee the safety and reduce miscommunications between the ATCOs and pilots. In this work, we propose some baselines on the SC task aimed at detecting the speaker role from transcribed ATC communications (sentences). Our previous work on speaker role detection is covered in~\cite{zuluaga2021bertraffic,prasad2021grammar}.

\subsubsection{Experimental Setup}

The SC experiments are carried out in a very related manner to NER (see Section~\ref{subsec:ner_es}). Specifically, we use the same model (\texttt{bert-base-uncased}), hyperparameters (e.g., number of epochs), optimizer, dropout rates and so on. However, here, we fine-tuned the model on the SC task rather than NER. We append a linear layer with a dimension of 4 (following the classes structure from Section 3.2 of~\cite{nigmatulina2022two}) on top of the last layer of the BERT model, i.e., a two-class classification model.

We employed LDC-ATCC\footnote{The Air Traffic Control Corpus (LDC-ATCC) corpus is public in: \url{https://catalog.ldc.upenn.edu/LDC94S14A}. It comprises recorded speech for use in the area of ASR for ATC. The audio data is composed of voice communication traffic between various controllers and pilots.} and 
UWB-ATCC\footnote{The UWB-ATCC corpus is released by the University of West Bohemia, and it can be downloaded for free in: \url{https://lindat.mff.cuni.cz/repository/xmlui/handle/11858/00-097C-0000-0001-CCA1-0}. The corpus contains recordings of communication between ATCOs and pilots. The speech is manually transcribed and labeled with the speaker information, i.e., whether ATCO or pilot is speaking and when.}
datasets (see Table~\ref{tab:databases}) for fine-tuning and \textit{ATCO2-test-set corpus} for testing. 
In LDC-ATCC and UWB-ATCC databases, speaker roles tags for each sample are marked in the original transcripts. 
And, we use ground truth ASR transcripts the evaluation.
We create speaker-independent train/test splits based on the original databases. The split IDs for each subset are registered in the public GitHub repository of this paper.

\subsubsection{Results}

We report the baseline results for speaker role detection in Table~\ref{tab:spk_id_results}. Differently from NER, we only used \textit{ATCO2-test-set corpus} for evaluation. We trained three models using different training datasets. From Table~\ref{tab:spk_id_results} we can see that pilots' communications are more challenging for our model in comparison to the ones from ATCOs. For instance, in the model fine-tuned with LDC-ATCC corpus, there is a two-point drop in F1-scores for pilots, i.e., 0.79 $\rightarrow$ 0.77 F1-score. Similar behavior is seen in the model fine-tuned with UWB-ATCC corpus, i.e., a four-point drop in F1-scores, 0.86 $\rightarrow$ 0.82. However, models trained on the later show more robustness for both classes in comparison to the one trained with LDC-ATCC.

We also investigated the performance benefit of combining both datasets. For this experiment, we only obtained one point increase for the pilot class, while one point decrease for the ATCO class, both in comparison to the model trained on UWB-ATCC only. It is important to keep in mind that \textit{ATCO2-test-set corpus} is a completely unseen dataset throughout all the experiments. We are convinced that integrating a small in-domain development set could boost the performances.

\begin{table}[t]
    \caption{Different performance metrics for the speaker role detection experiments. Metrics reported on \textit{ATCO2-test-set corpus} with a \texttt{bert-base-uncased} model trained on the splits from Table~\ref{tab:databases_experiments}. @P, @R and @F1, refers to precision, recall and F1-score, respectively. Numbers in \textbf{bold} refers to the top performance per column.}
    \label{tab:spk_id_results}
    \centering
    
    \begin{tabular}{lllllllllll}
        \toprule
        
        \multirow{2}{*}{\textbf{Training Corpus}} &  & \multicolumn{3}{c}{\cellcolor{Gray}\textbf{\textcolor{teal}{ATCO}}} &  & \multicolumn{3}{c}{\cellcolor{Gray}\textbf{\textcolor{magenta}{PILOT}}} &  & 
        \cellcolor{Gray}\textbf{AVG} \\
        \cline{3-5} \cline{7-9}
         & \multicolumn{1}{c}{} & \multicolumn{1}{c}{@P} & \multicolumn{1}{c}{@R} & \multicolumn{1}{c}{@F1} & \multicolumn{1}{c}{} & \multicolumn{1}{c}{@P} & \multicolumn{1}{c}{@R} & \multicolumn{1}{c}{@F1} & \multicolumn{1}{c}{} & \multicolumn{1}{c}{@F1} \\
        \midrule
         
        LDC-ATCC &  & 0.87 & 0.73 & 0.79 &  & 0.70 & 0.86 & 0.77 &  & 0.78 \\
        UWB-ATCC &  & 0.88 & \textbf{0.83} & \textbf{0.86} &  & \textbf{0.80} & 0.85 & 0.82 &  & \textbf{0.84} \\
        \midrule
        LDC-ATCC + UWB-ATCC &  & \textbf{0.92} & 0.78 & 0.85 &  & 0.76 & \textbf{0.91} & \textbf{0.83} &  & \textbf{0.84} \\
        \bottomrule
    \end{tabular}
\end{table}

\section{Legal and privacy aspects for collection of ATC recordings}
\label{sec:legal-ethics}

In order to safely distribute and make available the ATCO Corpus to the community, we took into account legal and ethical considerations as a prerequisite to distribute this content both, commercially and as open-source package.
The main question we faced was to determine whether we could legally record and distribute ATC. To answer that question, we inquired into how legislation and regulations treat ATC~\cite{rigault2022legal}.

Our first hypothesis was that ATC would fall under the rules of Intellectual Property law which regulate how authors and companies can collect and use immaterial works such as recordings, where we believed ATC could fall into. To confirm this hypothesis, we aimed to find out whether ATC could be considered as material that could be protected by intellectual property legislation. 
Therefore, we performed a thorough study of the two major legal intellectual property systems of the United States and Europe. This study showed that due to the specific characteristics of ATC, such as phraseology and context, these conversations could not be protected as such as they do not meet the originality threshold for protection. 

Then we moved on to another hypothesis. We thought that even if these conversations can not be protected as such, they might be protected as part of databases collected by either aircraft companies or Air Navigation Services Providers (ANSPs). As part of our investigations, we came into contact with some of these stakeholders, but none of them replied favorably to our requests. For example, the National Air Traffic Services which handles the airspace for the United Kingdom replied to us that they could not provide their recordings unless mandated by a Court order. Moreover, the United Kingdom is one of the few countries that expressly prohibit recording ATC communications, as its legislation strictly prohibits the use of unlicensed recording apparatus\footnote{Further information in the following url: \url{https://www.legislation.gov.uk/ukpga/2006/36/section/48}}. 

Other countries have a more lenient policy towards access and recording of ATC. Indeed, during our research we found out that ATC recorded in the US could be accessed on demand by formulating a request to the Federal Aviation Administration (FAA, hereafter). This is made possible by the use of the Freedom of Information Act (FOIA) that compels US administrations to make available certain types of information collected by these administrations during their operations. In the specific case of the FAA, Freedom of information regulations states that radio and computer data can be obtained upon request as stated in Chapter 4 Section 4-8-2 of the Facility Operation and Administration Order \footnote{JO 7210.3CC—Facility Operation and Administration available at \url{https://www.faa.gov/air_traffic/publications/atpubs/foa_html}}. Nevertheless, during our exchanges with the FAA we found out that requests for audio files needed to concern the last 45 days as required by Chapter 12, section 2 Article 12-2-2 of the same order and had to be specific to an airport in order to be processed adequately by the administration. Future work may include the drafting of such a request to confirm the reality of those conversations. 

For the airports based in Europe, we based our collection process on the existence of the Open Data Directive, formerly known as the Public Sector Directive, whose goal is to allow reuse and redistribution of data collected by public services, as we found out, many ANSPs are either State-owned or run by state administration for obvious security reasons. Therefore, we made a request in France to access data collected by the administration in charge of Air Traffic Control. We based our request on the provisions of French Law allowing to request the access to data produced by this administration. Regarding ATC, our request went up to the Commission d'Accès aux Documents Administratifs (Commission for access to administrative documents). This Commission ruled that the Direction Générale de l'Aviation Civile (DGAC) did not have to fulfill our request for data since they could not differentiate between civilian and military aircraft and that the recordings could leave the identification of speakers. However, following an \textit{a contrario} interpretation, it can be assumed that since our project focussed on civilian aircraft and that we ensured the anonymization of the conversations before making them available, we could pursue data collection. 

This previous ruling raised our concerns regarding the compliance of the project especially with the regulations related to protection of personal data, especially the EU General Data Protection Regulation (GDPR). GDPR is the main text regulating the collection and distribution of databases containing personal data. In the case of ATC, the speech data contains voiceprints of the pilots and ATCOs, which can be used as a mean of identification through speaker identification techniques. Thus, further precautions should be adopted in order to be able to collect this type of data. However, GDPR allow the collection of speech data when made in relation to reasons of substantive public interest, which we found applicable in our case since the project is aimed at enhancing airspace security.

\section{Conclusions and Future Work}

This article introduces, the \textit{ATCO2 corpus}, a set of three corpora for research on robust automatic speech recognition and natural language understanding of air traffic control communications. In ATCO2, we have successfully created and deployed an operating pipeline for collecting, pre-processing and automatically transcribing ATC audio data. During the data collection period, we mostly relied on a worldwide community of volunteers that acted as `data feeders'. Then, a community of `data annotators' employed the SpokenData transcription platform to generate gold annotations of a small portion of the collected data. The platform is up and running, and it is reachable on \url{https://www.spokendata.com/atco2}. 

The \textit{ATCO2 corpus} is divided in \textit{ATCO2-PL-set corpus} and \textit{ATCO2-test-set corpus}. The former contains more than 5000 hours of automatically transcribed ATC speech data, spanning more than ten airports in different continents (Table~\ref{tab:stats-databases}). While the latter, \textit{ATCO2-test-set corpus}, contains gold annotations of 4 hours of ATC speech. A subset called \textit{ATCO2-test-set-1h corpus} is offered for free in \url{https://www.atco2.org/data}. 
To the authors' knowledge, this is the first public release of a large-scale database for research in the area of air traffic control communications. 

In addition, we also cover baselines (our source code for data preparation and to replicate the NLU baselines will be stored in the following public GitHub repository \url{https://github.com/idiap/atco2-corpus}.) over three different key tasks in the area of ATC. The first one is related to robust ASR, while the next two are about to NLU of ATC communications. 

(1) We demonstrated that training an ASR system solely on \textit{ATCO2-PL-set corpus} reaches competitive WERs on both, public and private databases (see Table~\ref{tab:kaldi_result_alldata}). This is important because \textit{ATCO2-PL-set corpus} is purely composed of pseudo labels generated by ATCO2 project seed ASR system. This can be the starting point for many researchers and companies worldwide that would like to use our corpora for testing and training robust ASR systems for ATC. 

(2) We demonstrated that as much as 3000 utterances are needed to train and evaluate a BERT-based Named Entity Recognition system for ATC communications. This system is capable of detecting callsigns, commands, and values from the textual inputs. This NLU task is of special interest to the ATC community because this high-level information can be used to assist ATCOs in order to reduce their overall workload.

(3) Similarly, we developed a simple yet efficient BERT-based module that performs speaker role detection from textual inputs. 

Finally, we believe that the `lessons learned' in ATCO2 project and its recipe for collecting and pre-transcribing large-scale audio databases can be easily transferred to other applications, where data scarcity is a latent problem. 

\section*{Acknowledgements}

This paper introduces the \textit{ATCO2 Corpus} derived from a joint contribution from Clean Sky 2 Joint Undertaking (JU) and EU-H2020.

The work was fully supported by Clean Sky 2 Joint Undertaking (JU) and EU-H2020, under Grant Agreement No. 864702—ATCO2 (Automatic collection and processing of voice data from air-traffic communications).


\bibliographystyle{IEEEbib}
\bibliography{biblio}

\pagebreak
\appendix

\section{Automatic Transcription Engine}
\label{appendix:transcription-pipeline}

This appendix describes in details how we collected the audio and metadata that brought to live the \textit{ATCO2 corpus}. We mainly rely on the automatic transcription engine, described in more details in Section~\ref{subsec:data-processing-pipeline}. The automatic transcription engine is implemented as a scalable cloud service. It communicates with other services (or partners) using APIs. This service is designed to process large flows of data produced by data feeders.\footnote{Enthusiasts that act as 'feeders' of ATC speech and contextual ATC data (surveillance). See Section~\ref{subsec:feeders}.}

\begin{wrapfigure}{R}{0.6\linewidth}
    \begin{center}
        \includegraphics[width=0.95\linewidth]{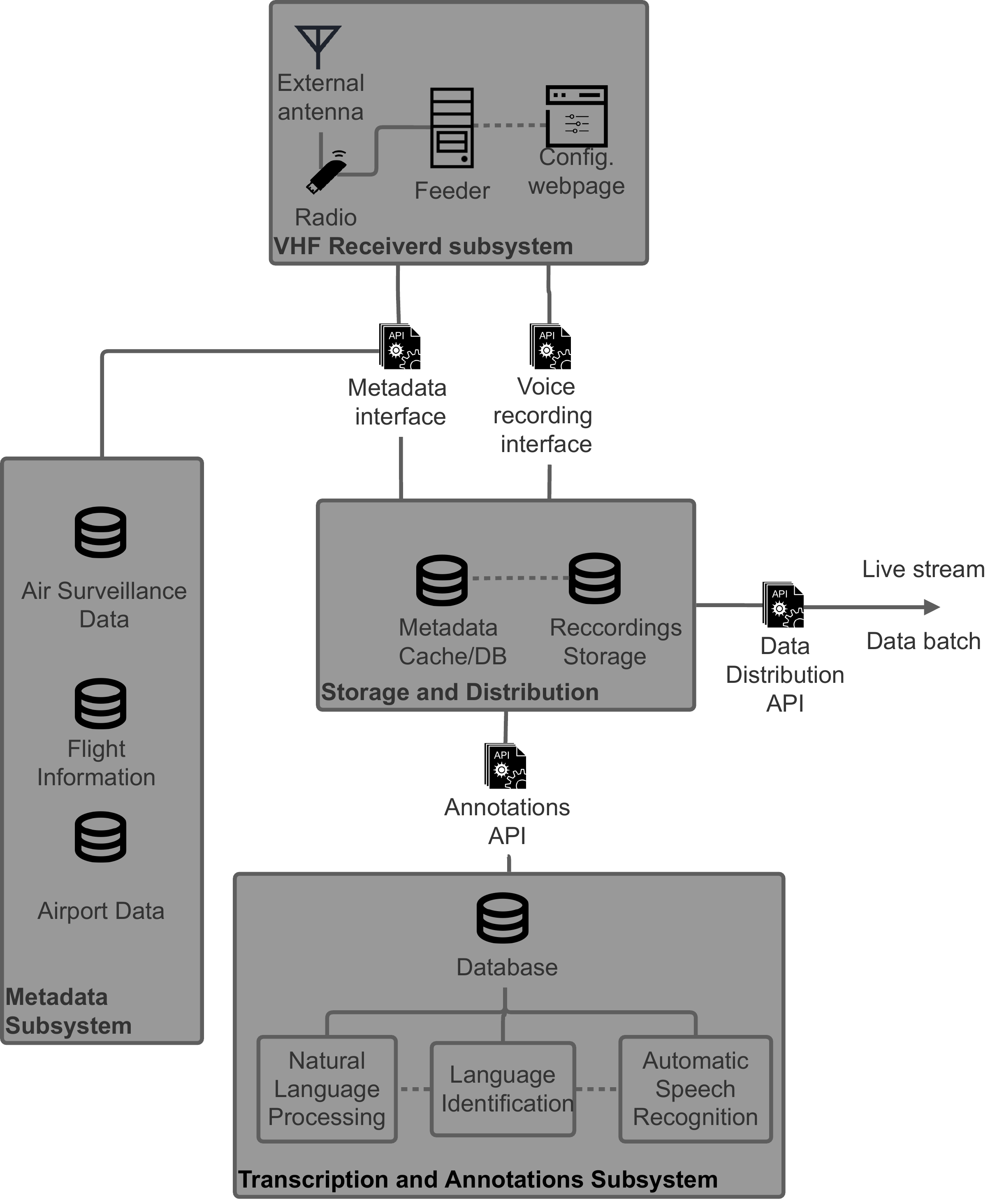}
        \caption{Overall ATCO2 communication schema.}
        \label{fig:full_collection_pipeline}
    \end{center}
\end{wrapfigure} 

The data is pushed to this service by OSN\footnote{OpenSky Network: \url{https://opensky-network.org/}.} servers by calling an API request and providing a job setting JSON file. After the request is accepted, settings parameters are processed and the job is stored in an internal queue for processing. The user (in this case, OSN) may have an ability to tweak the settings and to affect the processing pipeline and the result. Namely:

\begin{itemize}
    \item Audio input format choices;
    \item Rejection threshold for too long audios;
    \item Rejection threshold for too short audios;
    \item Rejection threshold for too noisy audios;
    \item Rejection threshold for non-English audios;
    \item Switching the language of automatic speech recognizer.
\end{itemize}

Most of these are actually disabled due to security reasons (not to interrupt the processing pipeline), but may be easily enabled on the fly if needed. The overall data flow model is described in Figure~\ref{fig:full_collection_pipeline}. 
Any new job (request for a full automatic annotation of recording) accepted via API on the SpokenData\footnote{Industrial partner: \url{https://www.spokendata.com/atco2}.} side is processed by a master processing node. The job is enqueued into a workload manager queue. Once there is a free processing slot, the job is submitted to a processing server, or worker. The master processing node then informs the OSN server about the state of the job by calling a callback.

\onecolumn
\pagebreak
\section{Unification of transcripts}
\label{appendix:unification}

This appendix conveys our main results of transcripts unification and lexicon formatting. Note that the description below is related to the databases employed to train the seed ASR engine used during the pre-transcription process, described in Section~\ref{subsec:data-processing-pipeline}. Special attention was devoted to the unification of words from the radiotelephony alphabet and numbers (see \textit{ICAO alphabet} column from Table~\ref{tab:appendix_unification}). Note that we map the word 'niner’ $\rightarrow$ 'nine’, and in the pronunciation lexicon, we allow the word 'nine’ to be pronounced as \verb|'n ay1 n er0’| (phoneme-based format).
Also, some standard expressions can be written as two words or as a single word. For some of the frequent ones, we selected one version that is used systematically (see \textit{Common expressions} column from Table~\ref{tab:appendix_unification}). We also rectify some airline designators that are part of the callsigns uttered by the ATCOs and pilots (see \textit{Airline designator} column from Table~\ref{tab:appendix_unification}).

We derive a table of mapping rules by extracting insights from a diverse list of airline designators. In total, we have a list of 5.4k airline designators, out of these, there are 1.8k multi-word airline designators. The airline designators ligatured by underscore are easier to be produced by the 'speech-to-text’ system as the tokens are longer, and there is also less uncertainty to be modelled by the language model. 
Finally, we pay extra attention to the transcripts generated by the ATCO2 community of volunteers. Like in any other human input, there might be typos or other types of transcription errors. It is necessary to at least revise the transcripts by the 2nd round of human inspection, where the errors are ideally fixed.

\begin{table}[h]
    \centering
    \caption{Normalization rules applied for annotation. }
    \label{tab:appendix_unification}
    \begin{tabular}{ l|l|l }
        \toprule
        \multicolumn{3}{c}{\cellcolor{Gray} \textbf{Unification of transcripts}} \\ 
        \midrule
        \textbf{ICAO alphabet} & \textbf{Common expressions} & \textbf{Airline designators} \\ 
        \midrule
        alpha $\rightarrow$ alfa & take off $\rightarrow$ takeoff & qatar $\rightarrow$ qatari \\
        charly $\rightarrow$ charlie & call sign $\rightarrow$ callsign & turkey $\rightarrow$ turkish \\
        juliet $\rightarrow$ juliett & readback $\rightarrow$ read back & air france $\rightarrow$ airfrans \\
        oskar $\rightarrow$ oscar & flightlevel $\rightarrow$ flight level & norshuttle $\rightarrow$ nor shuttle \\
        xray $\rightarrow$ x-ray & stand by $\rightarrow$ standby & airvan $\rightarrow$ air van \\
        zoulou $\rightarrow$ zulu & start up $\rightarrow$ startup & rynair $\rightarrow$ ryanair \\
        whisky $\rightarrow$ whiskey & goodbye $\rightarrow$ good bye & airbaltic $\rightarrow$ air\_baltic \\
        tripple $\rightarrow$ triple & clear for $\rightarrow$ cleared for & air berlin $\rightarrow$ air\_berlin \\
        niner $\rightarrow$ nine & lineup $\rightarrow$ line up & air canada $\rightarrow$ air\_canada \\
        0 $\rightarrow$  zero & clear for $\rightarrow$ cleared for & air china $\rightarrow$ air\_china \\
        1 $\rightarrow$  one & turnright $\rightarrow$ turn right & air europe $\rightarrow$ air\_europe \\
        2 $\rightarrow$  two & oclock $\rightarrow$ o'clock & jet stream $\rightarrow$ jet\_stream \\
        3 $\rightarrow$  three & o clock $\rightarrow$ o'clock & jetstream $\rightarrow$ jet\_stream \\
        4 $\rightarrow$  four & push back $\rightarrow$ pushback & k l m $\rightarrow$ k\_l\_m \\
        5 $\rightarrow$  five & descent direct $\rightarrow$ descend direct & klm $\rightarrow$ k\_l\_m \\
        6 $\rightarrow$  six & goodbye $\rightarrow$ good bye & korean air $\rightarrow$ korean\_air \\
        7 $\rightarrow$  seven & goodday $\rightarrow$ good day & koreanair $\rightarrow$ korean\_air \\
        8 $\rightarrow$  eight & turbulance $\rightarrow$ turbulence & wizzair $\rightarrow$ wizz\_air \\
        9 $\rightarrow$  nine & til $\rightarrow$ till & top\_jet $\rightarrow$ topjet \\ 
        \bottomrule
    \end{tabular}
\end{table}

\pagebreak
\section{How a Sample From ATCO2 corpora Looks Like?}
\label{appendix:sample-xml-tagged}

Example of human annotations for a recording of \textit{ATCO2-test-set corpus} in XML format. This file encodes most of the metadata. If more than one segment is present, it means there are two or more in the recording:

\begin{lstlisting}[language=XML, caption=XML tagged example from \textit{ATCO2-test-set corpus}. This example contains two recordings separated by the \textit{<segment>} tag.]
<?xml version="1.0" encoding="utf-8"?>
<data>
        <segment>
                <start>0</start>
                <end>2.93</end>
                <speaker>B</speaker>
                <speaker_label>pilot</speaker_label>
                <text>[unk] [#callsign]Quebec Lima[/#callsign] [#command]confirm cleared for ILS[/#command] [unk]</text>
                <tags>
                        <correct>0</correct>
                        <correct_transcript>1</correct_transcript>
                        <correct_tagging>0</correct_tagging>
                        <non_english>0</non_english>
                </tags>
        </segment>
        <segment>
                <start>2.99</start>
                <end>10.45</end>
                <speaker>A</speaker>
                <speaker_label>ATCO approach</speaker_label>
                <text>[unk] [#callsign]Quebec Lima[/#callsign] [#command]affirm cleared ILS approach[/#command] [#value]runway one four[/#value] [#command]if you go around[/#command] [#value]runway one four[/#value] [#command]report in localizer established[/#command]</text>
                <tags>
                        <correct>0</correct>
                        <correct_transcript>1</correct_transcript>
                        <correct_tagging>0</correct_tagging>
                        <non_english>0</non_english>
                </tags>
        </segment>
</data>
\end{lstlisting}

Basic details from the previous XML tagged segment: 

\begin{itemize}
    \item <segment> $\dots$ </segment>: one sample of data. One recording may have one or more segments;
    \item <start> $\dots$ </end>: timing information with speech activity by the speakers;
    \item <speaker> $\dots$ </speaker>: speaker information to identifiy whether the segment is from an ATCO or pilot. Unknown cases are tagged with <UNK>
    \item <text> $\dots$ </text>: ground truth transcripts with high-level entities annotations (callsigns, commands and values). Not all the segments contains these annotations.
    \item <tags> $\dots$ </tags>: extra metadata.
\end{itemize}

\end{document}